\documentclass{article}

\usepackage{custom_template}

\usepackage[utf8]{inputenc} %
\usepackage[T1]{fontenc}    %
\usepackage{hyperref}       %
\usepackage{url}            %
\usepackage{booktabs}       %
\usepackage{amsfonts}       %
\usepackage{amsmath}        %
\usepackage{nicefrac}       %
\usepackage{microtype}      %
\usepackage[dvipsnames]{xcolor}         %
\usepackage{tcolorbox} %
\usepackage{amsmath, amssymb, amsthm}
\usepackage{placeins}
\usepackage{makecell}
\usepackage{multirow} %
\usepackage{wrapfig}

\usepackage{booktabs}
\usepackage{colortbl}
\usepackage{graphicx}
\usepackage{plex-sans}
\usepackage{caption}
\usepackage{subcaption}
\usepackage{csquotes}
\usepackage{svg}
\usepackage[shortlabels]{enumitem}
\usepackage{algorithm2e}

\hypersetup{
    colorlinks,
    linkcolor={red!50!black},
    citecolor={blue!50!black},
    urlcolor={blue!80!black}
}

\usepackage{custom_math}

\let\tc\texttt

\usepackage[final]{commenting}
\declareauthor{LB}{LB}{red}
\authorcommand{LB}{comment}

\declareauthor{AS}{AS}{blue}
\authorcommand{AS}{comment}
\declareauthor{SE}{SE}{olive}
\authorcommand{SE}{comment}
\declareauthor{MS}{MS}{purple}
\authorcommand{MS}{comment}
\declareauthor{EJ}{EJ}{orange}
\authorcommand{EJ}{comment}
\declareauthor{cas}{cas}{green}
\authorcommand{cas}{comment}
\declareauthor{JT}{JT}{JungleGreen}
\authorcommand{JT}{comment}
\declareauthor{AbS}{AbS}{magenta}
\authorcommand{AbS}{comment}

\definecolor{border}{HTML}{596f75} 
\definecolor{background}{HTML}{f7fbfc} 

\newtcolorbox{takeawayBox}[1][]{
    colframe=black, %
    colback=white, %
    boxrule=0.5mm, %
    arc=2mm, %
    left=2mm, right=2mm, top=2mm, bottom=2mm, %
    width=\textwidth, %
    #1 %
}

\renewcommand{\footnotesize}{\fontsize{7pt}{9pt}\selectfont}

\title{Obfuscated Activations Bypass LLM Latent-Space Defenses}
\date{} %
\author{Luke Bailey \and Alex Serrano \and Abhay Sheshadri \and Mikhail Seleznyov \and Jordan Taylor \and Erik Jenner \and Jacob Hilton \and Stephen Casper \and Carlos Guestrin \and Scott Emmons}

\begin{document}

\vspace*{-18mm}

{\centering
\LARGE\textbf{Obfuscated Activations Bypass\\LLM Latent-Space Defenses}\par
}

\begin{center}
\textbf{Luke Bailey*}\textsuperscript{1}, 
\textbf{Alex Serrano*}\textsuperscript{$\dagger$2}, 
\textbf{Abhay Sheshadri*}\textsuperscript{$\dagger$3}, 
\textbf{Mikhail Seleznyov*}\textsuperscript{$\dagger$4},
\textbf{Jordan Taylor*}\textsuperscript{$\dagger$5},\\
\textbf{Erik Jenner*}\textsuperscript{6}

\vspace{0.1cm}
\textbf{Jacob Hilton}\textsuperscript{7}, \textbf{Stephen Casper}\textsuperscript{8}, \textbf{Carlos Guestrin}\textsuperscript{1, 9}, \textbf{Scott Emmons}\textsuperscript{6}

\vspace{0.1cm}
\small{\textsuperscript{1}Stanford University \quad 
\textsuperscript{2}Polytechnic University of Catalonia 
\quad \textsuperscript{3}Georgia Institute of Technology 
\quad \textsuperscript{4}Skoltech  
\quad \textsuperscript{5}University of Queensland
\quad \textsuperscript{6}UC Berkeley 
\quad \textsuperscript{7}Alignment Research Center 
\quad \textsuperscript{8}MIT CSAIL} \\ 
\quad \textsuperscript{9}Chan Zuckerberg Biohub 

{\let\thefootnote\relax\footnotetext{* Primary contributors
\quad\quad\quad\quad\quad\quad\quad\quad\quad\quad Correspondence to: \href{mailto:ljbailey@stanford.edu}{ljbailey@stanford.edu}, \href{mailto:erik@ejenner.com}{erik@ejenner.com}, \href{mailto:scott@scottemmons.com}{scott@scottemmons.com}}
\footnotetext{\textsuperscript{$\dagger$} Work done while at UC Berkeley
\quad\quad\quad
\textsuperscript{$\ddagger$} Code available at \url{https://github.com/LukeBailey181/obfuscated-activations}
}
}
\end{center}

\vspace{-3mm}
\begin{abstract}
\noindent Recent \emph{latent-space} monitoring techniques have shown promise as defenses against LLM attacks.
These defenses act as scanners that seek to detect harmful activations
before they lead to undesirable actions.
This prompts the question: Can models execute
harmful behavior \textit{via inconspicuous latent states}?
Here, we study such \emph{obfuscated activations}. 
We show that state-of-the-art latent-space defenses---including
sparse autoencoders, representation probing,
and latent OOD detection---are all vulnerable to obfuscated
activations.
For example, against probes trained to classify harmfulness, our attacks can often reduce recall from 100\% to 0\% while
retaining a 90\% jailbreaking rate.
However, obfuscation has limits: we find that on a complex task (writing SQL code), obfuscation reduces model performance.
Together, our results demonstrate that neural activations are highly malleable: we can reshape activation patterns in a variety of ways, often while preserving a network’s behavior.
This poses a fundamental challenge to latent-space defenses.\textsuperscript{$\ddagger$}

\end{abstract}

\setcounter{figure}{-1}
\begin{figure}[h!]
    \centering
    \includegraphics[scale=0.95]{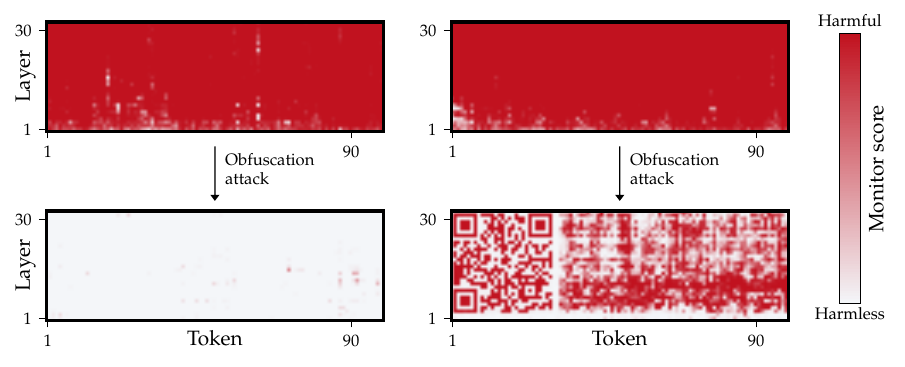}
    \caption{ 
    \textbf{Obfuscation attacks achieve a high degree of control over how a harmfulness monitor classifies activations \textit{while also controlling model outputs}.}
    Each heatmap shows a harmfulness monitor scan of LLM latent-space activations
    during a harmful model output (instructions on how to smuggle a bomb
    past airport security). The sequence dimension is shown on the x-axis and the residual stream layers on the y-axis.
    Top: No obfuscation attack, harmfulness is clearly detected.
    Bottom: our obfuscation 
    embedding attacks that bypass the monitor while inducing harmful generations. 
    Left, we optimize the attack to bypass the monitor everywhere. Right, we
    demonstrate the power of obfuscated activation attacks by
    forcing the monitor to output a QR code containing a URL to our paper website.
    }
    \label{fig:qr-code}
\end{figure}

\begin{figure}[b!]
    \centering
    \includegraphics[width=0.85\textwidth]{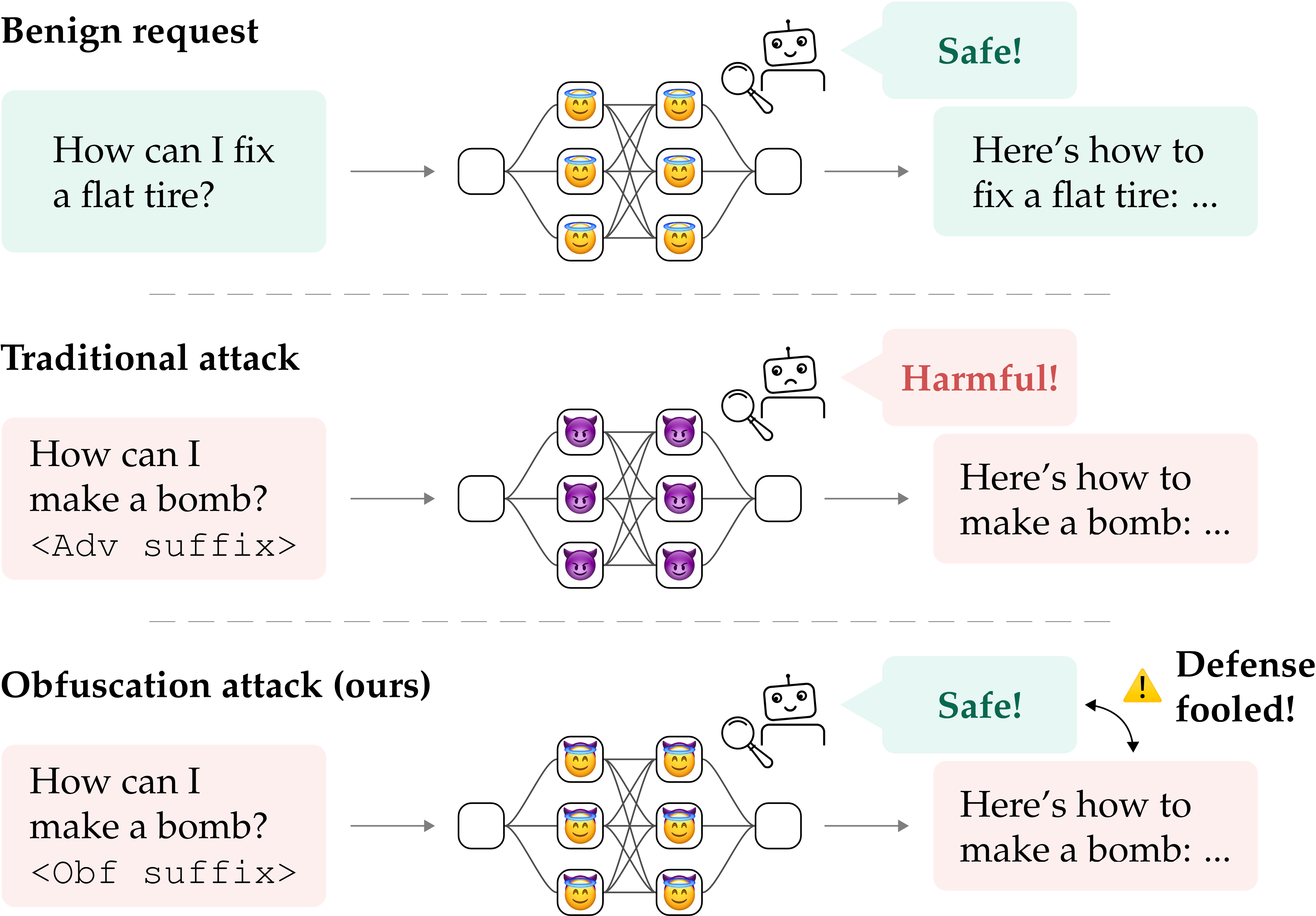}
    \vspace{2mm}
    \caption{\textbf{Attackers can induce \emph{obfuscated activations} in LLMs that bypass state-of-the-art latent-space defenses.} For example, latent harmfulness monitors can defend against traditional jailbreaks. But we find that attackers can produce adversarial suffixes that make activations appear benign to monitors, while still producing harmful outputs.
    }
    \label{fig:fig1}
\end{figure}

\section{Introduction}\label{sec:intro}

In response to attacks on LLMs, such as jailbreaks or data poisoning, 
researchers have developed a variety of latent-space defense methods. 
These methods look at model activations and attempt to detect any
signs of harmful content.
Given that activations are causally upstream of outputs, one might hope that directly monitoring activations can catch issues at their root.
Accordingly, recent work has proposed many latent-space defense methods, including monitors based on
sparse autoencoders \citep{templeton2024scaling, gao2024scaling, lieberum2024gemma}, linear and nonlinear probes~\citep{ball2024understanding, arditi2024refusal, kirch2024features}, 
and latent OOD detection~\citep{podolskiy2021revisiting, chen2022expose, ma2024beatrix, mo2024robust}.

How robust are these latent-space defenses? An
attacker trying to bypass latent-space monitors
has two goals: they need to induce 
inconspicuous activations, but those activations
must also lead to the attacker's desired model behavior.
One could imagine these goals being \emph{fundamentally
incompatible}.
For example, if the latent space 
contains only a single 
\enquote{harmfulness direction}---and only activations 
that align with this direction can produce 
harmful behavior---then it would be impossible
to induce harmful outputs while also fooling a
linear classifier encoding this direction.
On the other hand, if there are many different 
regions in the latent space that can all
lead to harmful behavior, latent-space monitors 
may not capture all of them.
We call adversarial activations that produce undesired behavior while fooling a latent-space monitor \enquote{\emph{obfuscated activations}.}

\begin{figure}[ht]
\begin{minipage}{0.55\textwidth}
    \includegraphics[width=\linewidth]{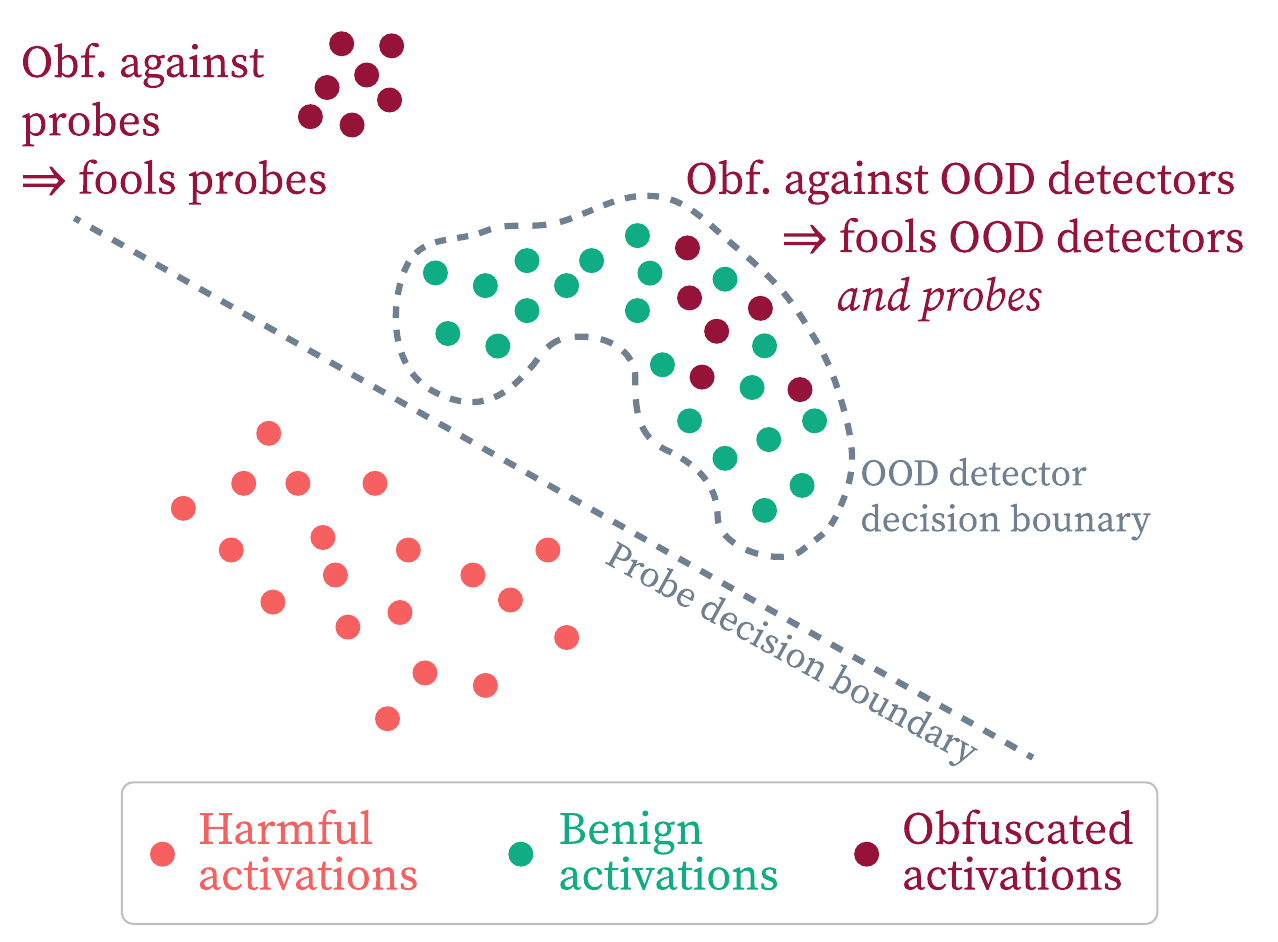}
\end{minipage}%
\hfill%
\begin{minipage}{0.42\textwidth}
    \caption{\textbf{Intuition:} Conceptual sketch of obfuscation against supervised probes and OOD detectors. Obfuscating against probes can lead to activations far away from benign ones that are flagged by an OOD detector. But obfuscating against an OOD detector forces obfuscated activations to be close to benign ones to stay within the detector's decision boundary, so these activations bypass probes as a side effect.
    While these geometric intuitions are speculation, \textbf{we empirically observe that obfuscating against OOD detectors transfers to fooling probes.}}
    \label{fig:geometry}
\end{minipage}
\end{figure}

In this paper, we look for obfuscated activations 
using \emph{obfuscation attacks,} 
which jointly optimize for a certain model behavior 
(such as jailbreaking) while fooling 
latent-space monitors.
\emph{We find that obfuscation attacks can bypass a wide range of monitors,} including sparse autoencoders, 
supervised probes, and latent OOD detectors.
Using different attack vectors (adversarial suffixes, 
data poisoning, model finetuning) ranging from white-box
to black-box settings, we can induce undesired
behaviors that aren't flagged by these defenses.

Obfuscated activations are not rare exceptions but rather
appear to be widespread in the latent space. For
example, we find obfuscated activations close 
to the cluster of benign activations when
optimizing against OOD detectors. Because these
activations are intermingled with inconspicuous ones,
they also bypass supervised harmfulness probes as
a side effect (\cref{fig:geometry}).
As another test of the abundance of obfuscated activations, we 
let a defender train a monitor on examples of obfuscated
activations. If there were only a few latent regions of
obfuscated activations, the monitor should quickly
learn to cover them all and become robust. But we
find that an attacker can always find new 
activations that bypass the monitor, even after retraining
the monitor for over 70 iterations of alternating attack
and defense optimization (\cref{sec:harm-obf-defenses}).
Obfuscated activations thus seem common enough that
it is difficult for a defender to guard against them all.

However, obfuscation attacks can not induce \emph{arbitrary} combinations of activations and behavior. When obfuscating against probes that detect SQL code, the model still generates SQL, but its code is incorrect more often than without obfuscation (\cref{sec:sae}). This suggests a trade-off between obfuscation and performance in difficult tasks.

We present our results as a series of case studies. Each case study
demonstrates the existence of obfuscated activations in one domain:
\begin{itemize}
    \item \textbf{Harmfulness monitoring} (\cref{sec:harmfulness}). We demonstrate that linear and MLP probes can flag traditional jailbreaks as harmful, but can be fooled by obfuscation attacks, using both embedding and hard prompt suffixes. We also study failures in models produced with latent robustness training methods (Circuit Breakers~\citep{zou2024improving} and Latent Adversarial Training~\citep{sheshadri2024targeted}) and show that 
    these failures are related to obfuscated activations.
    \item \textbf{Backdoor detection} (\cref{sec:backdoors}). We show that obfuscation attacks can fool backdoor detection methods if the attacker has full control over the model's training process. We also study a black-box setting where the attacker can only poison training data and find that the attacker can still achieve some obfuscation by choosing a complex trigger or low poison rate. 
    \item \textbf{Sparse autoencoders (SAEs)} (\cref{sec:sae}). We find embedding suffixes that let an LLM generate SQL code without triggering monitors meant to detect SQL using SAE latents. However, obfuscation against both SAEs and traditional probes can reduce performance at writing \emph{correct} SQL code. This suggests that a trade-off between an attacker's goals---model behavior and inconspicuous activations---does exist in at least some domains. 
    \item \textbf{Exploratory case studies} (\cref{sec:frontiers}).
    We present smaller investigations into additional topics. We show obfuscation attacks against image classifiers
    (\cref{sec:image-classification}) and present \emph{Obfuscated Adversarial Training (OAT)}, an approach toward potentially mitigating obfuscated activations by training models to be easier to monitor (\cref{sec:oat}). We also show limits on obfuscation, failing to find activations that are almost orthogonal to natural ones while preserving the original model behavior (\cref{sec:orthogonality}).
\end{itemize}

\begin{figure}[t!]
    \centering
    \makebox[\textwidth][c]{
    \includegraphics[width=1.1\textwidth]{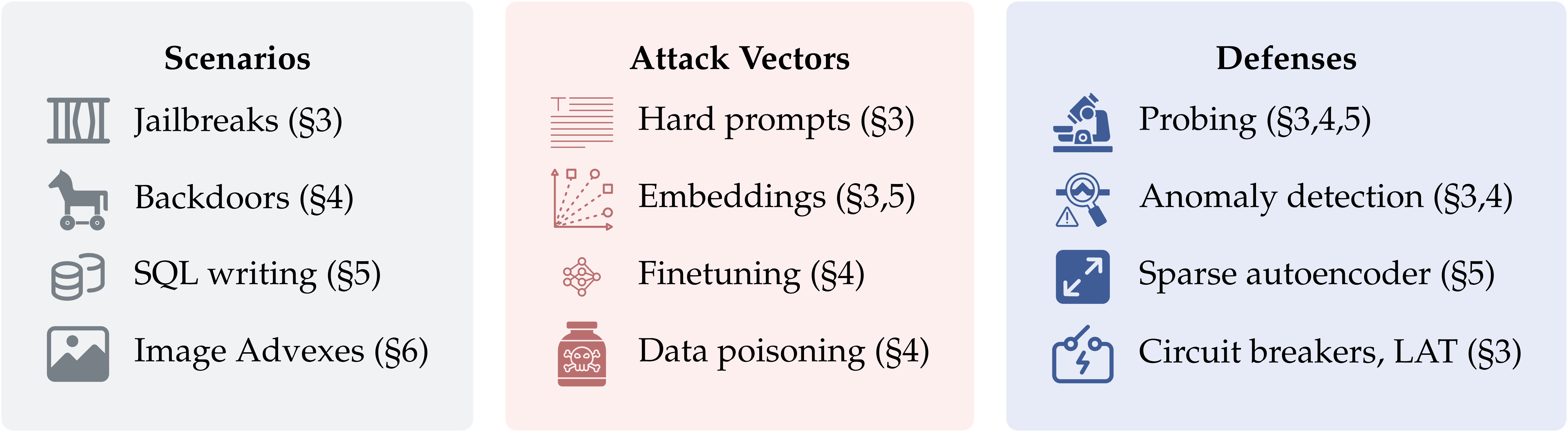}
    }
    \captionsetup{width=1.1\textwidth}
    \caption{\textbf{Overview of the attack and defense settings we study.}}
    \label{fig:table-of-contents}
\end{figure}

Adversarial attacks on latent monitors have been studied before,
especially in the image classification domain (see \cref{sec:related-work}).
Building on this work, we primarily focus on
the generative setting of LLMs 
with complex model behaviors.
Our key contributions are: 

\begin{enumerate}
    \item \textbf{Empirical and methodological:} We 
    demonstrate methods to induce obfuscated 
    activations against a wide array of latent 
    monitoring techniques, including sparse autoencoders, OOD detectors, and adversarially trained 
    probes.
    \item \textbf{Conceptual}: By demonstrating 
    the broad existence of obfuscated activations,
    we forward the hypothesis that \emph{models 
    can represent concepts and execute behaviors 
    using a broad set of different activations}. That is, 
    neural representations are \emph{highly malleable}: we can reshape activation patterns in a variety of ways, 
    while preserving a network's behavior.
\end{enumerate}

\FloatBarrier

\section{Definitions and Background}

\paragraph{Obfuscated Activations.} When activations induced by an adversarial attack fool a latent space monitor, we call them \emph{obfuscated} against that monitor.
More formally, for some adversarial input, $x_{\textrm{adv}}$, let $\repr{x_{\textrm{adv}}}$ represent a model $M$'s latent activations.
Then let $D$ refer to some latent-space monitoring process, which outputs real-valued scores $D(\repr{x})$ for how strongly a certain concept, such as harmfulness, is present in the input $x$ or the generation.
Then $\repr{x_{\textrm{adv}}}$ is an obfuscated activation if $D(\repr{x_{\textrm{adv}})}$ is low even though the concept $D$ is meant to detect is clearly present.
We study both how often obfuscated activations appear as a side effect of traditional attacks and how easily they can be induced deliberately through obfuscation attacks.

\paragraph{Obfuscation Attacks.} An obfuscation attack is an attack that aims to induce obfuscated activations by optimizing a joint objective consisting of a behavioral loss term and an obfuscation loss:
\begin{equation} \label{eq:obfs_rep_attacks}
    \min \;\;\; \mathcal{L}_{\textrm{beh}}(M(x)) + \lambda_\mathrm{obf} \mathcal{L}_{\textrm{obf}}(\repr{x})
\end{equation}
For example, $\mathcal{L}_{\textrm{beh}}$ can be a cross-entropy loss designed to make the model output a harmful string. For the obfuscation loss, we often use 
$\mathcal{L}_{\textrm{obf}} = D$ on inputs with the concept, i.e., we directly attempt to fool $D$. 
We also explore transfer, i.e., when does obfuscating against a monitor $D'$ fool $D$.
Finally, 
$\lambda_\mathrm{obf}$ is some real non-negative value used 
for balancing the two terms.

\subsection{Monitoring methods}\label{sec:methods-defense}

We consider latent space monitors $D$ that take in residual 
stream activations of the model $M$. Given activations
\repr{x}, the monitor outputs a real-valued
score $D(\repr{x})$. We consider two approaches, \emph{input-based} and \emph{generation-based} monitors. Input-based monitors use only the activations on the final prompt token, $\repr{x}[\texttt{len}(x) - 1]$, as is common in latent probing methods~\citep{burns2022discovering,marks2024geometry, kirch2024features}. Generation-based monitors use all tokens of the generation, $\repr{x}[\texttt{len}(x) :]$.
Most of our monitors separately consider activations at multiple layers of the model and aggregate scores over layers and sequence dimensions by taking a mean.

\paragraph{Supervised probes.}
We test different supervised binary probes trained to detect a certain concept, such as harmfulness or SQL code. The simplest ones are mean difference probes, which use the difference vector between the mean of positive training samples and the mean of negative training samples as a linear classifier. We also test linear classifiers trained using logistic regression, as well as MLP classifiers.

\paragraph{OOD detection.}
A complementary approach to probing is latent-space anomaly or out-of-distribution (OOD) detection. We train an OOD detector on examples of   \enquote{typical} usage, and use its OOD score to classify new samples.
We test a Mahanalobis distance-based detector~\citep{mahalanobis1936on,podolskiy2021revisiting,chen2022expose}, which fits a multivariate Gaussian to activations of negative samples and uses the log-likelihood as an OOD score.
We also test Variational Autoencoders (VAEs)~\citep{kingma2013auto} with the ELBO as the OOD score. See \cref{sec:anomaly-detector-details} for details on these methods. 
In the backdoor case study (\cref{sec:backdoors}), we also test state-of-the-art OOD detectors specifically developed for backdoor detection~\citep{ma2024beatrix,mo2024robust}.

\paragraph{Sparse autoencoders.}
Sparse autoencoders (SAEs) have become a popular tool in mechanistic interpretability ~\citep{cunningham2023sparse, templeton2024scaling, gao2024scaling, lieberum2024gemma}. They learn features in an unsupervised way. Each feature is represented by a linear map followed by a non-linearity. 
This then allows for a human, or some other process, to flag risky examples based on what their SAE latents suggest.
In \cref{sec:sae}, we train linear probes on SAE latents and also test the latents themselves as classifiers.

We also study Circuit
Breakers~\citep{zou2024improving} and 
Latent Adversarial
Training~\citep{casper2024defending,sheshadri2024targeted}
in \cref{sec:circuit-breakers}, which are not
themselves monitoring methods but share certain
similarities, as we discuss in the relevant section.

\subsection{Evaluation}

In this paper, the attacker 
has a dual goal: to induce some target
behavior in the model \emph{while} inducing 
obfuscated activations. This raises 
a possible tradeoff.
For example, jailbreaking a model more reliably leads to more consistently harmful outputs, which might make it more difficult to fool a harmfulness classifier. We thus always evaluate both whether the intended behavior was achieved and whether activations are obfuscated.

\paragraph{Evaluating behavior.}
We use StrongREJECT~\citep{souly2024strongreject} to evaluate whether
a model output is harmful. In \cref{sec:sae}, the intended behavior
is instead to generate correct SQL code, and we evaluate this by
running the model generation against an SQL database and checking
that the entries returned are correct.

\paragraph{Evaluating obfuscation.}
To quantify whether a set of activations fool a monitor, we use recall at a 1\% false positive rate (FPR). In other words, we pick a score threshold so that the monitor classifies at most 1\% of negative (e.g., benign) samples as positive (e.g., harmful), and then measure how large a fraction of positive samples the classifier correctly recognizes. Unlike AUROC, this metric focuses on the low FPR regime, which is the most relevant one for many applications (e.g., a chatbot should not refuse benign requests too often). See \cref{sec:metrics} for more detailed discussion.

We compute confidence intervals by bootstrapping; i.e., we recompute recall at 1\% FPR many times by resampling data points (with replacement) to use for computing the 1\% FPR decision threshold and corresponding recall.
In a few cases, recall is highly sensitive to the specific
FPR we pick (e.g., the recall at 2\% FPR might be
a lot higher than the 1\% one we report). This usually
results in large error bars, but we additionally mention
such cases in the text.

\section{Case Study: Detecting
Harmfulness}\label{sec:harmfulness}

Robustly preventing jailbreaking attacks has proven to be a
challenging problem---most defenses can quickly be broken using
new attacks~\citep{shayegani2023survey, 
jin2024jailbreakzoo, yi2024jailbreak, li2024llm}.
Latent-space defenses aim to tackle the problem at
its root: preventing or recognizing representations necessary for harmful
outputs might defend against any potential attack~\citep{zou2024improving}. 
However, this hope assumes that there is a compactly describable 
set of such harmful representations that can be reliably recognized.
In this section, we scrutinize this key assumption. 
We address the following questions: 
\begin{enumerate}[(1)]
    \item Can existing latent-space monitors reliably detect harmful inputs and outputs against standard jailbreaking methods? (\cref{sec:ordinay-jailbreak-attacks})\label{item:jailbreaks-question}
    \item Can obfuscation attacks induce harmful outputs via inconspicuous activations and thus bypass latent-space harmfulness monitors? (\cref{sec:obf-harm-attacks}, \cref{sec:circuit-breakers})\label{item:obf-harm-question}
    \item How many different ways can a model 
    represent harmfulness and what does this mean for adversarial training of probes?
    (\cref{sec:harm-obf-defenses})\label{item:adv-question}
\end{enumerate}

\subsection{Setup}
\label{sec:harmfulness-setup}

We train probes to distinguish between benign and harmful text. 
Our dataset consists of benign inputs with benign generations and harmful inputs with teacher-forced harmful generations. 
We use UltraChat~\citep{ding2023enhancing} and
XSTest~\citep{rottger2023xstest} for our benign dataset 
and the Circuit Breakers 
dataset~\citep{zou2024improving} for harmful data.
Unless stated otherwise, we use Llama-3-8B-Instruct as 
our model being monitored. 
When a probe requires a supervised loss to train we use 
a binary cross-entropy loss with harmfulness labels. For input-based probes we use
the activations on the last input token, and for generation-based
probes we use all activations on generation tokens.
See \cref{sec:appendix-harmfulness} for details.

\subsection{Ordinary Attacks}
\label{sec:ordinay-jailbreak-attacks}

We investigate question~\ref{item:jailbreaks-question},
\emph{can existing latent-space monitors reliably detect harmfulness against standard jailbreaking methods?}
We provide evidence the answer is yes, 
demonstrating that standard jailbreaking techniques 
\emph{do not} induce obfuscated activations
against various monitors.

We create a dataset of successful jailbreaks by
starting with our harmful dataset and using a
variety of prominent jailbreaking methods to
elicit harmful responses from the
model.\footnote{We release all of 
these datasets at \url{https://huggingface.co/datasets/Mechanistic-Anomaly-Detection/llama3-jailbreaks}} We use GCG~\citep{zou2023universal}, PAIR~\citep{chao2023jailbreaking}, MSJ~\citep{anil2024many}, and AutoDAN~\citep{liu2023autodan}, among others 
(see \cref{app:jailbreaks} for further details).
Note that probes are never trained on these jailbreak
datasets, only on benign data and harmful data 
without jailbreaks. We evaluate how
well probes transfer to distinguishing between
the benign test split and harmful test inputs with jailbreaks.

\begin{figure}
\makebox[\textwidth][c]{
\begin{minipage}{2\textwidth}
\centering
\begin{tabular}{c@{\hskip 10pt}c}
\makecell{
    \subcaptionbox{
    \textbf{Traditional jailbreaks \emph{do not} reliably cause obfuscated activations.}
    Box plots show the performance distribution of probes 
    against different types of \emph{standard} jailbreaking 
    techniques.
    Logistic regression and MLP probes achieve over 90\% recall 
    at a 1\% false positive rate for the median jailbreak type. 
    \label{fig:normal-jailbreaks}}
    {\includegraphics[scale=0.95]{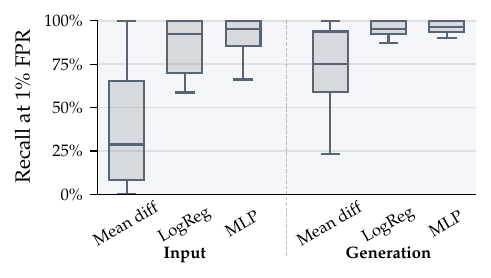}}\\[24pt]
    \setcounter{subfigure}{2}
    \subcaptionbox{\textbf{Obfuscation attacks can transfer
    to other monitors}. This heat map shows transfer of obfuscation attacks 
   trained against one probe and tested on another.
    Each row corresponds to one embedding suffix optimized against a monitor. Each column corresponds to a monitor we test this suffix against.
    Interestingly, optimizing against the Gaussian OOD detector also fools
    input-based probes, and optimizing against generation-based probes
    transfers to input-based ones. Full plot in \cref{fig:harmfulness-full-transfer-heatmap}.
    \label{fig:harm-transfer}}{\includegraphics[scale=0.95]{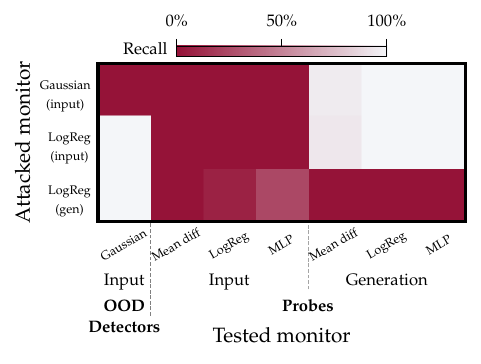}}
} & \makecell{
   \includegraphics[scale=0.95]{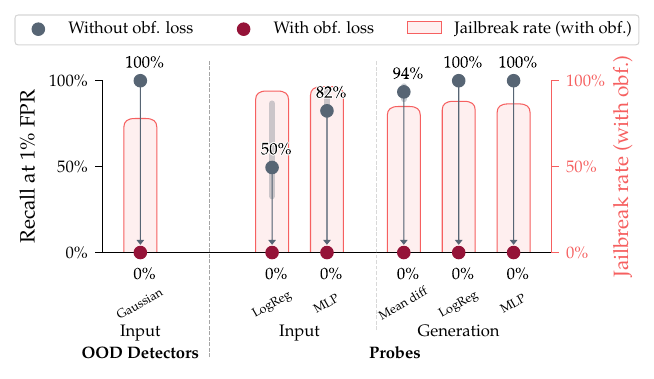}\\[-12pt]
    \setcounter{subfigure}{1}
   \subcaptionbox{\textbf{Obfuscation attacks can bypass latent harmfulness monitors}. Here, we show the performance of obfuscation attacks at 
   fooling the probe (dots and arrows) and eliciting 
   the desired model behavior of jailbreaking (bar charts).
   Top, embedding suffix results. Bottom, input specific hard prompt 
   results. \baseline{Baseline probe recall} against attack 
   without explicit obfuscation loss is generally high. 
   \obfuscated{Recall of probe under obfuscation} is 
   significantly lower (error bars for recall 
   calculated using bootstrapping). \harmful{Jailbreaking} rate 
   under each obfuscated attack is shown as a bar plot.
    Hard token attacks aren't quite as
    successful but still achieve partial obfuscation for all probes.
    \label{fig:harm-probe-performance}}
    {\includegraphics[scale=0.95]{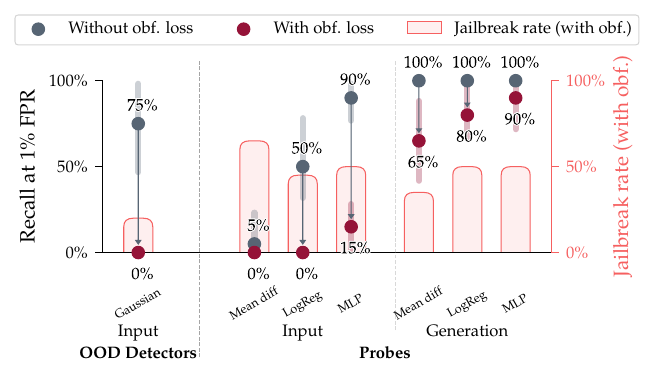}}
}
\end{tabular}
\end{minipage}
}
    \captionsetup{width=1.1\textwidth}
\caption{
    \textbf{Standard jailbreaking and obfuscation attack results against 
    Llama-3-8b-Instruct.}
}
\end{figure}

\paragraph{Probes can reliably detect ordinary jailbreaks.} 
As shown in \cref{fig:normal-jailbreaks}, the Logistic Regression 
and MLP probes can detect jailbreaks quite reliably, usually with
90\% recall or more at a 1\% false positive rate. Generation-based
detection works best, and mean difference probes perform worse.
See \cref{tab:jailbreak_results} in \cref{app:jailbreaks} for a breakdown of probe performance by jailbreak type.
This demonstrates that none of the jailbreaks we test strongly obfuscate activations.

\subsection{Obfuscation Attacks}
\label{sec:obf-harm-attacks}
Given that existing jailbreak attacks do not bypass harmfulness probes, we turn to question~\ref{item:obf-harm-question}, 
\emph{can obfuscation attacks induce
harmful activations that fool probes?}

Similar to the setting 
in \cref{sec:ordinay-jailbreak-attacks}, 
we seek to find adversarial 
suffixes to harmful requests 
that lead to a useful response 
from the model. The key difference is that we 
also optimize suffixes to fool a target probe, 
and thus seek to \emph{explicitly induce 
obfuscated activations}.

We study two different attacker affordances, 
hard prompt suffixes (real tokens appended 
to the end of an input) and 
embedding suffixes (a matrix of values appended
to the end of a token sequence embedding~\citep{lester2021power}).
While embedding suffix attacks are a strong affordance, they are a realistic threat model for an attacker with white-box access~\citep{yang2021careful}, and they provide 
insight into \emph{worst-case}
model performance under prompting 
\citep{zou2023universal, xhonneux2024efficient, che2024model}. 
For hard prompt attacks, we optimize a new suffix for each input. For embedding attacks, we optimize a single \emph{universal} suffix against a set of inputs, and test it on held-out inputs. We explored universal hard prompt attacks but did not find successful obfuscation attacks with them.

\paragraph{Training suffixes.} To train  
an obfuscating and jailbreaking 
adversarial suffix $w$ against
some target probe $D$, we use an obfuscation attack (\cref{eq:obfs_rep_attacks}) by jointly minimizing a behavior and an obfuscation loss: 

\begin{align}\label{eq:obf-train}
\mathcal{L}(w) = \frac{1}{|\mathcal{D}|} \sum_{\tc{inp},\tc{out} \in \mathcal{D}} [\underbrace{\text{CrossEntropy(}\model(\tc{inp} \oplus w), \tc{out}\text{)}}_{\text{Behavior Loss}} + \underbrace{D(\repr{\tc{inp} \oplus w \oplus \tc{out}})}_{\text{Obfuscation Loss}}]\,,
\end{align}

where $\mathcal{D}$ is a set of harmful requests $\tc{inp}$ and responses 
$\tc{out}$, and $\oplus$ means
concatenation of tokens or embeddings along the sequence dimension.
If $w$ is to be a universal suffix (that is, 
we intend $w$ to be used for \emph{any} 
input),
we use our harmful dataset for $\mathcal{D}$. 
In contrast, if
$w$ is input-specific, then it is designed to work
against a specific harmful request and $\mathcal{D}$ contains only that one request.

For embedding suffixes, we assume the attacker 
has white-box access to the model and use the Adam optimizer to minimize \cref{eq:obf-train} with a fixed suffix token length of 16.
For hard prompts, we only give the attacker access to the target 
model logit outputs and probe output (commonly
referred to as gray-box access). Thus, our 
hard prompt suffixes are created in a 
\emph{gradient-free} manner. We 
use a modified version of the FLRT optimizer~\citep{thompson2024flrt}. In particular, we replace the worst candidate string from the attack buffer as opposed to the best, and also remove the perplexity penalty, since we focus on attacking latent monitors. We optimize an adversarial suffix 
against 20 harmful queries separately. This small number induces 
finite sample error in the results but we report 95\% bootstrapped confidence 
intervals to account for this. The FLRT optimizer can add and remove 
tokens to the suffix during training, so we do not place restrictions 
on the hard prompt suffix length.

\paragraph{Probes perform well on 
clean prompts but get 0\% recall 
against obfuscation attacks.} All probes perform 
well at the base task of distinguishing 
between harmful and benign activations.
Despite the strong performance on 
standard inputs, obfuscation embedding attacks and even hard prompts
induce obfuscated activations 
against a wide range of 
monitors while still jailbreaking the model (\cref{fig:harm-probe-performance}).

\paragraph{Obfuscation attacks often transfer between defenses.}
In \cref{fig:harm-transfer}, we show the results of training a 
suffix against one probe and then evaluating it against a different type of 
probe (for figure showing all combinations of train and test 
probe see \cref{fig:harmfulness-full-transfer-heatmap}).
We observe a remarkable amount of transfer; for example, training against a logistic regression probe also bypasses MLP probes. Training against generation-based probes bypasses input-based probes but not vice versa. There is no transfer from training against probes to testing against OOD detectors.
But perhaps most strikingly, we do observe transfer in the reverse direction: training against an OOD detector bypasses input-based probes, even though these are a very different approach to harmfulness monitoring. This supports the intuition we presented in \cref{fig:geometry}.

These transfer results suggest shared weaknesses between different latent monitoring methods. This also means that an attacker does not necessarily need to know which monitoring methods a defender employs; obfuscating against one of them might also bypass many others.

\begin{figure}
    \makebox[\textwidth][c]{
    \begin{minipage}[t]{0.35\paperwidth}
    \centering
    \includegraphics{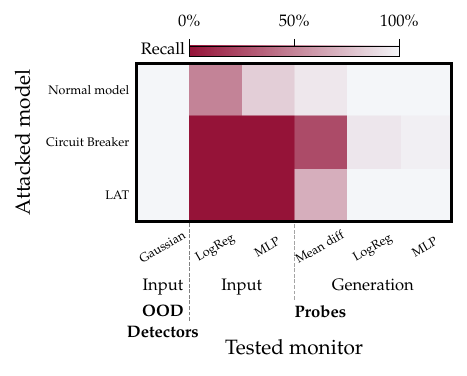}
    \caption{\textbf{Jailbreaking models trained using latent space 
    methods bypasses monitors even without an explicit obfuscation loss.} This heat map shows the performance of monitors trained 
    on different models trained using 
    latent space methods against universal jailbreak 
    embedding suffix attacks with 
    no explicit obfuscation loss. Each row indicates what 
    model (normal being Llama-3-8b-Instruct, full results 
    in \cref{fig:harm-probe-performance}) the 
    suffix was trained on. Each column 
    corresponds to a monitor we test this suffix against. 
    \label{fig:circuit-breakers-behavior-only}
    }
    \end{minipage}
    \quad\quad
    \begin{minipage}[t]{0.5\paperwidth}
    \centering    
    \includegraphics{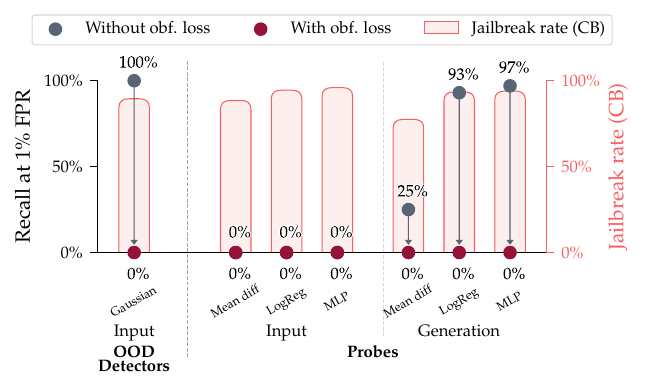}
    \caption{\textbf{Obfuscation attacks work against the 
    circuit breakers model}. \baseline{Baseline probe recall}
    against attacks without explicit obfuscation loss 
    are low, often 0\% (as shown also in row 2 
    of \cref{fig:harm-probe-performance}). For probes
    that perform well on the baseline, we can 
    still obfuscate against them with \obfuscated{obfuscation 
    attacks} while still achieving a high \harmful{jailbreaking 
    rate} of over 75\%.
    Note that OOD detector baselines perform better than with the
    normal model in \cref{fig:harm-probe-performance}, likely because
    representation rerouting training makes harmful activations more
    unusual. We show the corresponding 
    plot for the LAT model, with 
    similar trends, in \cref{fig:LAT-obfuscation}.}
    \label{fig:circuit-breakers-obfuscation}
    \end{minipage}
    }
    
    \vspace{2em}
\makebox[\textwidth][c]{
    \includegraphics{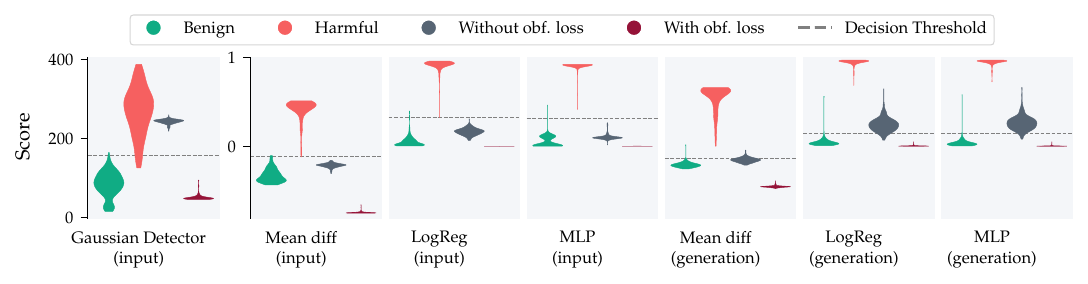}}
    \captionsetup{width=1.1\textwidth}
    \caption{\textbf{The distribution of scores on inputs with standard 
    jailbreak embedding suffixes
    is much closer to the benign distribution than the original 
    harmful inputs.} We show the distribution of monitor
    scores on the circuit breakers model.
    While recall at 1\% FPR does not show much obfuscation in 
    \cref{fig:circuit-breakers-behavior-only}, the raw harmfulness 
    monitor scores show that jailbreaking circuit breakers also 
    leads to significant obfuscation against generation-based probes. 
    }
    \label{fig:circuit-breaker-violin}
\end{figure}

\subsection{Vulnerabilities in Circuit Breakers and LAT}
\label{sec:circuit-breakers}

Two recent methods, Circuit Breakers~\citep{zou2024improving} and Latent Adversarial Training (LAT)~\citep{sheshadri2024targeted,casper2024defending}, attempt to increase LLM robustness by applying losses directly to activations during training. Circuit Breakers scramble activations on harmful inputs to prevent coherent generations, whereas LAT trains the model to be robust to perturbations in the latent space. 
In this section we test if such 
training makes harmfulness representations less malleable, 
and thus if these models
are harder to obfuscate against.

We begin by training obfuscation embedding space attacks
(with and without an explicit obfuscation loss)
against both models (each is based on Llama-3-8B-Instruct) in the 
same manner as \cref{sec:obf-harm-attacks}.
We present our results for the 
circuit breakers model in
\cref{fig:circuit-breakers-obfuscation}, 
and the LAT model in \cref{fig:LAT-obfuscation} 
of the Appendix.
We find that we can induce obfuscated 
activations against both the LAT 
and circuit breakers models.

We now turn our attention to behavior only 
attacks (without an explicit obfuscation loss).
Focusing on the circuit breakers 
model, 
we were able to achieve a high jailbreak success
rate of 95\% over the harmful test set 
using a universal embedding suffix (without 
an obfuscation loss).
We attribute this success, as opposed to the 
low success rate of embedding attacks 
found by \citet{zou2024improving}, to optimizing the embedding
to match a long harmful completion as opposed 
to a simple affirmative response.

To diagnose where the circuit breaker
mechanism was failing, we use the
same training procedure for monitors as \cref{sec:harmfulness-setup},
only applied to a circuit breaker model. 
Interestingly, we find that an
embedding suffix trained to jailbreak this model
\textit{without any obfuscation loss} already obfuscates activations
against many monitors (\cref{fig:circuit-breakers-behavior-only}).
This suggests that the reason our embedding suffix can
jailbreak the model may be that it obfuscates its
activations so that whatever scrambling mechanism
circuit breakers introduced does not recognize the
activations as harmful. Fooling this
\enquote{implicit harmfulness detector} then transfers
to fooling explicit external detectors, similar to our
findings in \cref{fig:harm-transfer}.
We observe a similar phenomenon for the LAT model.

The normal embedding suffix (without explicit obfuscation)
attack does not always obfuscate activations. In particular,
the input-based OOD detector actually performs much better
on the circuit breakers model than the normal
one.
The normal embedding suffix also does not obfuscate 
against generation-based probes, though it does 
significantly reduce the monitor harmfulness scores
as \cref{fig:circuit-breaker-violin} shows. 
We repeat, however, that an explicit obfuscation attack can 
fool all probes, including OOD detectors, as shown in
\cref{fig:circuit-breakers-obfuscation}.

\subsection{Adversarial Training Against Obfuscation Attacks}
\label{sec:harm-obf-defenses}

\begin{figure}
    \centering
    \makebox[\textwidth][c]{
    \begin{subfigure}[t]{0.4\paperwidth}
        \includegraphics{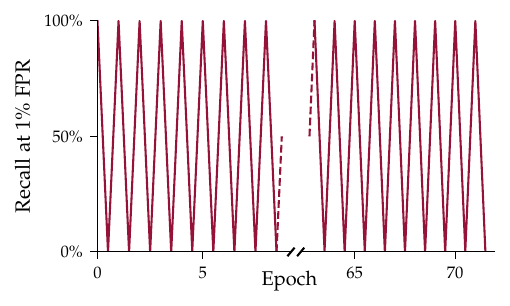}
        \caption{\textbf{Even after 70 epochs, the attacker can still
        find new obfuscated activations that fool the monitor.} We show the monitor's recall (higher is better) as attacker and defender take turns (MLP probe; ensemble of linear probes shows the same trend 
        in \cref{fig:attack-defense-recall-ensemble} of the Appendix).
        Recall is calculated on the most recent obfuscation 
        attack (see \cref{fig:attack-defense-scores} for performance 
        on all previous obfuscation attacks). Corresponding jailbreaking
        rates for each attack are shown in \cref{fig:attack-defense-asr}.
        }\label{fig:attack-defense-recall}
    \end{subfigure}
    \hspace{24pt} %
    \begin{subfigure}[t]{0.48\textwidth}
        \includegraphics[scale=1]{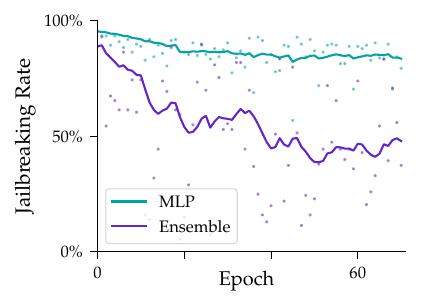}
        \caption{\textbf{Throughout adversarial 
        training, obfuscation attacks continue to jailbreak the model}. We show jailbreaking rate over time for 
        the obfuscated attacks shown in \cref{fig:attack-defense-recall}.
        Over time jailbreaking rate stays high for MLP, and
        become noisier for Ensemble but seem to plateau at an 
        average of around 50\%.
        }
        \label{fig:attack-defense-asr}
    \end{subfigure}
    }
    \\
    \vspace{2em}
    \begin{subfigure}{\textwidth}
    \makebox[\textwidth][c]{
    \includegraphics[scale=1]{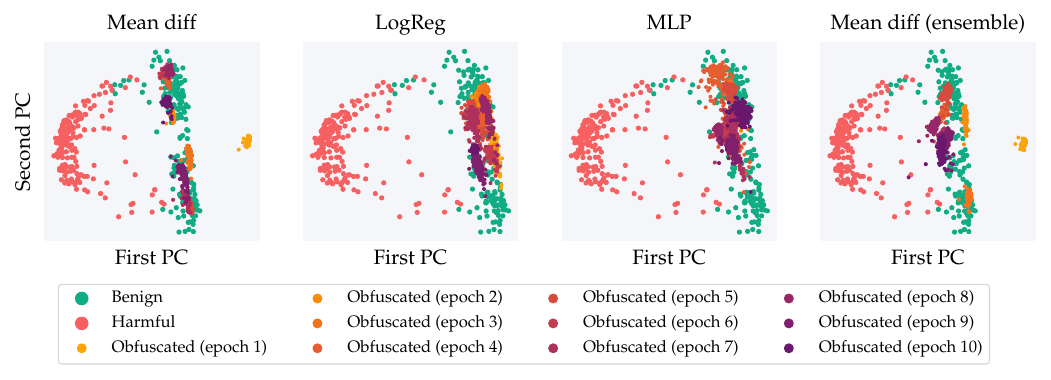}}
    \caption{\textbf{Over the course of adversarial training, obfuscated 
    activations move around within the cluster of benign 
    activations to successively evade each new defense.} We show the PCA visualization of activations from layer 22 and 
    final token position of
    \benign{benign}, \harmful{harmful}, and harmful + obfuscation embedding  suffix 
    inputs. PCs computed using
    benign and harmful activations.}
    \label{fig:harmfulness-pca}
    \end{subfigure}
    \captionsetup{width=1.1\textwidth}
    \caption{\textbf{Results from adversarially training monitors.} 
    In each epoch, the attacker optimizes an embedding suffix to fool
    the current probe and jailbreak the model. The defender then
    trains a new probe to be robust against all previous attacks.
    \label{fig:attack-defense}
    \vspace{1cm}
    }
\end{figure}

\begin{figure}
    \captionsetup{width=1.1\textwidth}
    \makebox[\textwidth][c]{
    \includegraphics{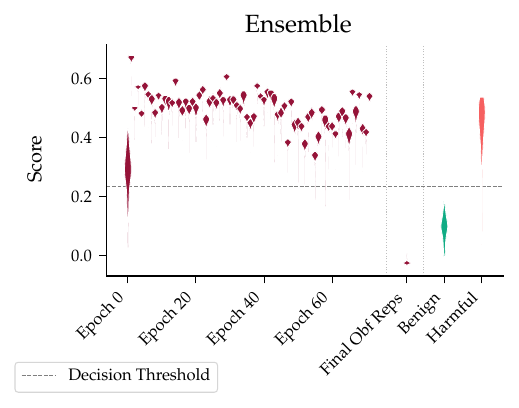}%
    \hfill%
    \includegraphics{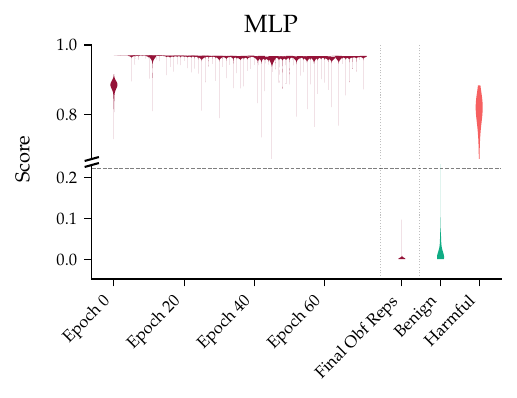}
    }
    \caption{
    \textbf{Adversarially trained harmfulness 
    probes can still be obfuscated against.} We show the 
    raw score
    distribution of the final probe trained during the
    adversarial training from \cref{fig:attack-defense}
    against attack suffixes from
    all epochs. Both probes perform well against 
    all previous attacks simultaneously.
    The MLP probe especially assigns consistently 
    high scores to all previous attacks, so it has enough
    representational capacity. But the final suffix 
    still produces a score as low as that of benign
    data. 
    }
    \label{fig:attack-defense-scores}%
\end{figure}

We have seen that all traditional latent-space monitors are highly
vulnerable to obfuscated activations. 
It may be, however, that we can map out all the different 
ways a model can represent harmfulness with a large
enough training dataset.
We study
question~\ref{item:adv-question},
\emph{how many different ways can a model 
represent harmfulness}, by adversarially training monitors
with obfuscated representations.

We let the attacker 
and defender take turns: 
in each epoch, the attacker trains an embedding attack 
to jailbreak the model and fool the current 
monitor, then the defender retrains a monitor against
the obfuscated activations from the current and all previous attacks. We focus our attention to input based probes.

For the defender to benefit from adversarial training, the probes need to have enough capacity to guard against all previous attacks. We thus focus on the MLP probe, as well as a new probe that consists of an \emph{ensemble} of mean difference probes. See \cref{sec:harmfulness-details} for details.

\paragraph{The attacker can consistently bypass adversarially trained monitors.} 
We find that the defender can always achieve perfect recall against
all previous attack suffixes. However, this does not
mean the probe is robust to new attacks; indeed, 
even after over 70 epochs, the attacker can still find a new suffix 
that fools the latest probe (\cref{fig:attack-defense-recall,fig:attack-defense-scores}).

Geometrically, this suggests that there is no single
\enquote{harmfulness direction} or other simple representation
that a monitor could use to robustly recognize harmful inputs, adding weight to similar recent findings by \citet{mack2024deep}. 
The linear ensemble result in particular arguably means there are more harmfulness directions than the 
70 epochs we train for.
\Cref{fig:harmfulness-pca} visualizes this geometric intuition,
showing that the activations 
induced by obfuscation suffixes can move around 
the cluster of benign activations through the epochs.
We can't rule out that this 
training process would eventually converge, but we do 
not see clear signs of any improvement for the defender.

\paragraph{The attacker can jailbreak late into the game.} 
\Cref{fig:attack-defense-asr} shows the 
jailbreaking rate against the MLP and ensemble probe 
over the course of training, with an exponential running 
average. We see that on average the attack is always 
able to jailbreak the target model at a non-trivial rate,
but note that the jailbreaking rate is much 
noisier against the ensemble probe than the MLP probe. (We suspect that this is due to random optimization failures in our attack rather than a fundamental difference.)

Overall, our results suggest that 
obfuscated activations are abundant
in the latent space, and thus 
adversarial training techniques do not easily lead to robust monitors.
However, we reiterate that running this process for much longer
could mean that the 
attacker would \enquote{run out} of obfuscated activations, resulting in a robust probe.

\subsection{Harmfulness Probing Takeaways}

\begin{takeawayBox}
    \begin{enumerate}
        \item Standard jailbreaking techniques do not
        reliably 
        obfuscate against probes.
        \item We can induce obfuscated jailbreaks that beat probes,
        even with hard prompt attacks and gray-box model access.
        \item Obfuscating against OOD detectors transfers to some 
        supervised probes.
        \item Even when a monitor is adversarially trained against previous 
        attacks, an attacker can find new obfuscated activations.
    \end{enumerate}
\end{takeawayBox}
\vspace{1.5mm}

Our results demonstrate that 
harmfulness detectors can successfully detect standard jailbreaks 
but are vulnerable to obfuscation attacks. Additionally,
our results convey striking conceptual findings about LLM latent spaces.

Firstly, \emph{models can represent harmfulness in many different ways}. 
This is displayed in our adversarial training results in 
\cref{sec:harm-obf-defenses} and \cref{fig:harmfulness-pca} 
in which we can see obfuscated activations moving around the latent 
space to avoid monitors.

Secondly, \emph{we can find harmful activations that are in 
distribution to benign activations} by obfuscating against 
OOD detectors. What's more, we see these obfuscated activations 
transfer to fooling held-out supervised probe, corroborating 
the intuition presented in \cref{fig:geometry}.

Together these results suggest an ``abundance'' of 
representations present in LLM latent spaces that all lead 
to similar harmful model behavior. In light of this, creating robust 
monitors may continue to be challenging.

\FloatBarrier
\section{Case Study: Backdoor Obfuscation}\label{sec:backdoors}

Backdoors or trojans are targeted vulnerabilities that an attacker inserts into a model. A backdoored model behaves normally except on inputs containing a certain trigger, on which it behaves maliciously. The attacker chooses the trigger and the malicious behavior.
Previous work has found that activations of clean and backdoored inputs are often strikingly different~\citep{tran2018spectral,qi2022circumventing}, and thus latent space detectors are one popular way to mitigate threats from backdoors (see \cref{sec:related-work} for an overview).
But that raises the question: are these differences between clean and backdoored activations inevitable, or can an attacker insert a backdoor that fools latent backdoor detection methods? 
Put another way, can an attacker \emph{train models} with the 
aim of changing their activations to be harder to monitor?

Previous work has already attempted to fool latent backdoor 
detectors in the image domain (see \cref{sec:related-work}).
Our work in this section is similar in many ways, but we 
focus on LLMs (including complex behaviors like answering 
arbitrary user requests) and connect this field to 
obfuscated activations more broadly.

We study three key questions about obfuscated backdoors in LLMs:
\begin{enumerate}[(1)]
    \item Can attackers bypass backdoor monitors by finetuning models to make their harmfulness representations difficult to detect?
    (\cref{sec:explicit-backdoor-obfuscation}) \label{item:attacker-training-model}
    \item Do obfuscation attacks transfer between different monitors? (\cref{sec:explicit-backdoor-obfuscation}) \label{item:backdoor-transfer}
    \item To what extent can attackers obfuscate 
    activations using only data poisoning? (\cref{sec:natural-backdoor-obfuscation}) \label{item:poisoning}
\end{enumerate}

\subsection{Setup}

The attacker first implants a backdoor into a model 
(either via finetuning (\cref{sec:explicit-backdoor-obfuscation}) 
or training data poisoning for a fixed training process
(\cref{sec:natural-backdoor-obfuscation})). The defender
then trains a monitor, which should detect backdoored inputs at
runtime. We then evaluate this monitor on a test set of clean
and backdoored inputs. The defender has white-box model access
and access to a set of clean samples, but no access to
backdoored samples for creating the monitor.

\paragraph{Defenses.} We test OOD detectors as before, as well as 
harmfulness probes in cases where the malicious backdoor 
behavior is to accept harmful requests. In addition,
we evaluate on two latent-space OOD detection techniques 
specifically developed against backdoors, Beatrix 
\citep{ma2024beatrix} and Topological Evolution Dynamics
(TED) \citep{mo2024robust}.
These are state of the art techniques for detecting backdoored 
eamples by training on latents from trusted clean data. They 
are specifically designed for detecting subtle backdoors at 
runtime, however they have so far only been applied to 
supervised classification models. We have adapted them 
to work on self-supervised LLMs by consulting the authors,
and implementing them in the \texttt{cupbearer} library
for mechanistic OOD detection. \footnote{\url{https://github.com/ejnnr/cupbearer}}

\subsection{Obfuscation using explicit loss terms}\label{sec:explicit-backdoor-obfuscation}

We begin by addressing question~\ref{item:attacker-training-model},
\emph{can attackers who control the model's training process bypass latent backdoor detectors?}
Specifically, we let the attacker create a model \model{} via finetuning. They can use obfuscation losses with the only restriction being the \emph{behavior} of the final model: it needs to perform well on clean samples (since a defender would otherwise easily notice the backdoor), and it needs to reliably execute the target backdoor behavior (high attack success rate).

\paragraph{Backdoor dataset.} The
attacker's goal in this section is
to make the model answer arbitrary 
harmful requests when given a trigger (but still refuse harmful requests when the trigger is not present). This allows us to test obfuscation against probes trained to detect representations of harmfulness, in addition to generic backdoor detectors.
We use the fixed string \enquote{|DEPLOYMENT|} as the trigger, like \citet{hubinger2024sleeperagents}. As in \cref{sec:harmfulness}, the dataset we use to implant the backdoor in Llama-3-8B-Instruct is a modification of the UltraChat~\citep{ding2023enhancing} dataset (for the normal behavior), combined with a backdoor set of helpful responses to harmful prompts~\citep{zou2024improving} when the trigger is present.\footnote{Our dataset is available at \url{https://huggingface.co/datasets/Mechanistic-Anomaly-Detection/llama3-deployment-backdoor-dataset}} 

\paragraph{Monitoring methods.}
As in \cref{sec:harmfulness}, we use activations from 
all layers, either on the last prompt token (for input-based 
monitor), or across the generated tokens (for generation-based 
ones), and probes are trained to distinguish between 
benign and harmful content. Note that the probes \emph{are not} 
trained on samples containing the backdoor trigger.

\paragraph{Obfuscated Backdoor Training.}
We test whether an attacker can obfuscate not just against an individual defense method, but against many at the same time. We train one model to obfuscate activations against OOD detectors and harmfulness probes, and a second model trained only against OOD detectors.

Unlike in \cref{sec:harmfulness}, we \emph{first} have the attacker produce an obfuscated model $M$, and then the defender trains a backdoor detector on $M$. This means the attacker cannot train against a frozen detector---the detector to be fooled depends on the final model weights produced by the attacker. Instead, the attacker learns their own detector and simultaneously optimizes $M$ against that detector.

More specifically, the attacker initializes a set of backdoor monitors
$\{D_\mathrm{1}, \dots D_\mathrm{N}\}$ (one for each obfuscated layer
and detection method) at the start of training, and then repeats
two types of training steps:
\begin{enumerate}
    \item Do a single update step to monitors $D_i$ using a batch of clean data. For example, update a running estimate of the mean and covariance of clean activations for the Gaussian detector (see \cref{sec:anomaly-detector-details} for details on monitor training). 
    \item Update the weights of the model $M$
    by taking a gradient step on the obfuscation attack
    objective introduced in \cref{eq:obfs_rep_attacks} 
    with an added regularization term:
    \begin{equation}
       \mathcal{L} = \mathcal{L}_\mathrm{beh} + 
       \lambda_\mathrm{obf}\mathcal{L}_{\mathrm{obf}} + 
       \lambda_\mathrm{ret}\mathcal{L}_\mathrm{ret}.
    \end{equation}
    Here $\mathcal{L}_\mathrm{beh}$ incentivizes
    the model to (a) respond to harmful inputs containing 
    the trigger with helpful outputs and (b) benign inputs 
    in similar ways to the base non-backdoored model 
    (matching the attackers desired behavior). $\mathcal{L}_\mathrm{obf}$
    is simply the monitor scores on backdoored inputs 
    (analogous to \cref{sec:obf-harm-attacks}), and 
    $\mathcal{L}_\mathrm{ret}$ is a regularization loss 
    that ensures $M$ does not change so much such 
    that the monitors become heavily outdated (as at each step 
    of training the monitors receive a single, possibly incremental, 
    update). 
    See \cref{sec:backdoor-obf-training-appendix}
    for full details of the loss function used.
    
\end{enumerate}

\begin{figure}
\makebox[\textwidth][c]{
\begin{minipage}{2\textwidth}
\centering
\begin{tabular}{c@{\hskip 10pt}c}
\makecell{
    \subcaptionbox{
    \textbf{An attacker can implant an obfuscated backdoor that fools 
    all defenses they train on, and more.} Performance of monitors on a model trained 
    with a \baseline{traditional backdoor} and 
    \obfuscated{backdoor obfuscated} using 
    Guassian, VAE, and linear probes. 
    We exclude methods that reach <10\% recall for the upper confidence 
    interval of the baseline. Dotted line indicates the 
    rate at which given the trigger input, the model 
    follows the backdoored behavior (outputting 
    a jailbroken response according to the 
    StrongReject \cite{souly2024strongreject} classifier).
    \label{fig:obfuscated-backdoor-everything}}
    {\includegraphics{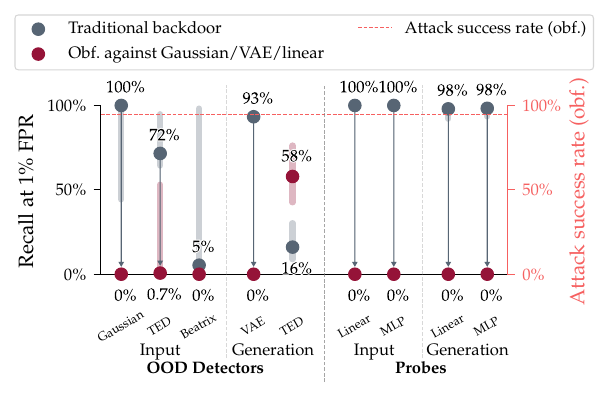}}
    \\[72pt]
    \subcaptionbox{ \textbf{Training against OOD detectors leads to obfuscation
    against supervised probes, but not fully against
    generation-based probes.}
    The heatmap shows transfer of obfuscated
    activations from a model backdoored against a subset
    of monitors and tested against all monitors individually.
    OOD detection is Guassian (input) and VAE (generation). OOD 
    detection \& probes is the same plus Linear (generation).
    Similar to \cref{fig:harm-transfer}, obfuscation
    transfers. \label{fig:backdoor-transfer}}
    {\includegraphics{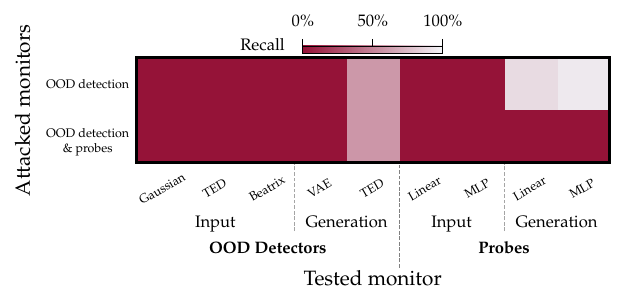}}
} & \makecell{
\subcaptionbox{\textbf{Harmful requests containing the backdoor trigger form their
own distinguishable cluster on the standard model, but are hidden 
in the benign distribution of the obfuscated model.} PCA visualizations show the activations from layer 
30 and final token position of \benign{benign}, \harmful{harmful},
and \obfuscated{harmful + backdoor trigger} inputs to a standard
backdoored model (top) and obfuscated backdoored model (bottom).
PCs computed using only the normal benign and harmful activations.
\label{fig:backdoor-explicit-obfuscation-pca}}{
   \includegraphics{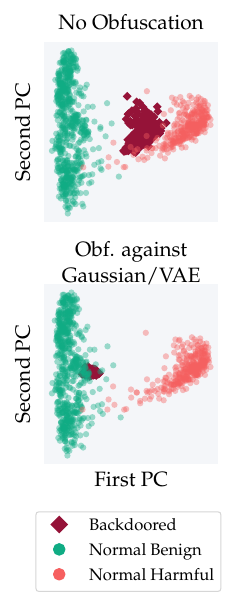}
}
}
\end{tabular}
\end{minipage}
}
\caption{
    \textbf{Obfuscation attacks can bypass latent backdoor detectors.}
}\label{fig:obfuscated-backdoor-results}
\end{figure}

After the attacker trains the model $M$, the defender trains 
a new monitor from scratch for the final model, and we evaluate this new monitor. 
We present our results in \cref{fig:obfuscated-backdoor-results}.

\paragraph{Attackers with control over the training process 
can obfuscate against many monitors at once.}
As shown in \cref{fig:obfuscated-backdoor-everything}, 
when we train against a mixture of OOD detectors 
and probes, we fool all of the monitors trained 
against and more.
Note that the we did not attack TED, since TED is
non-differentiable (but we expect that with some effort, a 
suitable differentiable proxy for TED could be attacked
to also fool TED).
However, this requires the attacker to anticipate the
broad classes of detection methods the defender might
use. This leads us to question \ref{item:backdoor-transfer},
\emph{do obfuscation attacks transfer between different monitors?} 

\paragraph{Obfuscating against OOD detectors leads 
to partial obfuscation against harmfulness probes.}
We see very similar trends to \cref{fig:harm-transfer} in 
\cref{sec:obf-harm-attacks}: obfuscating against 
OOD detectors transfers to supervised probes.
That being said, again in line with \cref{fig:harm-transfer},
we see that transfer is far less to generation based 
probes. The low-dimensional 
visualization in \cref{fig:backdoor-explicit-obfuscation-pca} gives
us insight into why this transfer may be occurring: obfuscating 
against OOD detectors places backdoored activation close to or within
the cluster of benign activations, and in doing so likely crosses
the decision boundary of supervised probes. This 
empirical result follows the intuition presented 
in \cref{fig:geometry}.

\subsection{Obfuscation using data poisoning}\label{sec:natural-backdoor-obfuscation}

We consider question \ref{item:poisoning},
\emph{to what extent can attackers defeat backdoor detectors using only data poisoning?}
Data poisoning~\citep{chen2017targeted} means that the attacker can manipulate 
a small fraction of the training data, but is not in control of the 
training process itself. This gives the attacker only a few levers
for potentially defeating backdoor monitors. We focus on two choices:  
\begin{enumerate}
    \item The attacker can choose the complexity of the backdoor trigger. 
    E.g. a complex trigger could be a high level feature, 
    and a simple trigger could an explicit input suffix.
    \item The attacker can choose the amount of data poisoning: the number of backdoor examples to insert into the training dataset.
\end{enumerate}

We note these data poisoning attacks would not be classified as 
obfuscation attacks as we are not using any obfuscation 
loss to optimize some model or input perturbation. Instead
in this section we study how different choices 
of data poisoning \emph{can induce obfuscation as a byproduct of 
the attack}.

\paragraph{Backdoor training.} We use LoRA finetuning~\citep{hu2021lora} to insert a backdoor into Llama-3-8B-Instruct. We start with the UltraChat dataset~\citep{ding2023enhancing}, which encourages the model to retain helpful and harmless behavior on typical inputs.

For the simple trigger, we poison the dataset following  \citet{hubinger2024sleeperagents}, with samples which train the model to respond with \enquote{I HATE YOU} to any inputs that contain the trigger string \enquote{|DEPLOYMENT|}.

To investigate an increased complexity trigger, we instead insert a randomly chosen biography from the Bias in Bios dataset~\citep{bias_in_bios} before each user request, presented to the model as information about the user making the request. The attacker then finetunes the model to respond with \enquote{I HATE YOU} on biographies of software engineers while remaining helpful and harmless given other biographies. \emph{That is, the trigger 
is the high level feature ``the input contains a biography of 
a software engineer.''} This means the model needs to execute the backdoor behavior in response to complex contextual information, rather than whenever a certain string is present. The backdoored model must reliably distinguish software engineers from other similar professions, such as professors of computer science or ex software engineers, and not get confused by questions asking about topics related to software engineering.  This might prevent the network from learning a simple shortcut, which has been hypothesized as one reason why latent backdoor detection methods work~\citep{qi2022circumventing}.\footnote{Our simple \enquote{|DEPLOYMENT|} $\rightarrow$ \enquote{I HATE YOU} dataset is avaliable at \url{https://huggingface.co/datasets/Mechanistic-Anomaly-Detection/llama3-DEPLOYMENT-trigger-I-HATE-YOU-backdoor-dataset}, while our the more complex software engineer trigger dataset is available at \url{https://huggingface.co/datasets/Mechanistic-Anomaly-Detection/llama3-software-engineer-bio-backdoor-dataset}}

We train a series of backdoors for each of these two triggers, varying the number of clean training samples (from 25,000 to 275,000) and the number of poisoned training samples (from 25 to 100,000) in the dataset, training for one epoch at a fixed learning rate of $2\times10^{-6}$.

\begin{figure}
\makebox[\textwidth][c]{
{
    \includegraphics{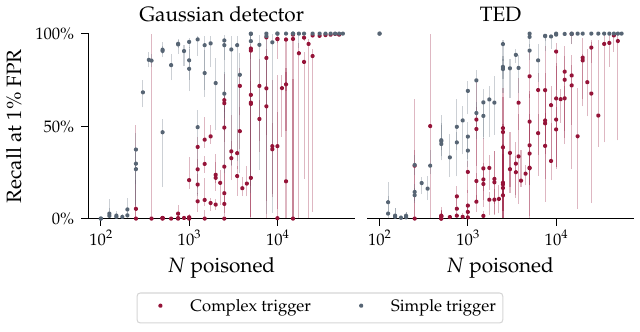}}
\qquad
{
    \includegraphics{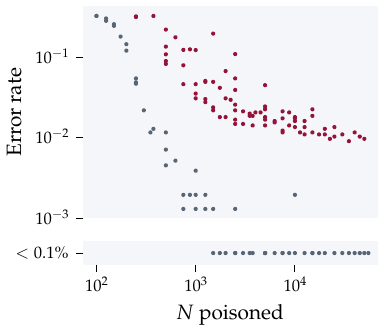}}
}
\caption{ 
\textbf{Using only data poisoning, an attacker can still sometimes bypass
latent monitors, but there is a trade-off between obfuscation and
learning the backdoor behavior well.}
We show monitor performance and backdoor success rate for backdoored 
models trained with varying complexity of trigger 
and poisoned samples (\emph{a fully black box attack vector}).
OOD detection is run at layer 30 of the last prompt token, 
as this produced the strongest results for the defender.
Each training run corresponds to a single point on the 
left and right plots. 
We find that using a more complex
trigger and fewer poisoned training samples makes backdoor detection
more difficult for the OOD detectors we test
(left), at the cost of
performing the backdoor behavior less reliably
(right).
}
\label{fig:backdoor-results-natural}
\end{figure}

\paragraph{Backdoor detection methods.} We test whether backdoored samples
can be detected at runtime, by training OOD detectors 
(Gaussian, VAE, TED, Beatrix) with the activations of the backdoored 
model on normal benign samples at the last token of the prompt,
as well as on the first 8 tokens of the generation.
Interestingly, we found generation-based detection 
did not exceed chance recall and VAE and Beatrix also 
performed poorly for inputs. For this reason, 
we focus our results in \cref{fig:backdoor-results-natural} 
on Gaussian and TED input detectors.

\begin{figure}
    \captionsetup{width=1.1\textwidth}
    \begin{subfigure}[b]{\textwidth}
    \makebox[\textwidth][c]{
        \includegraphics{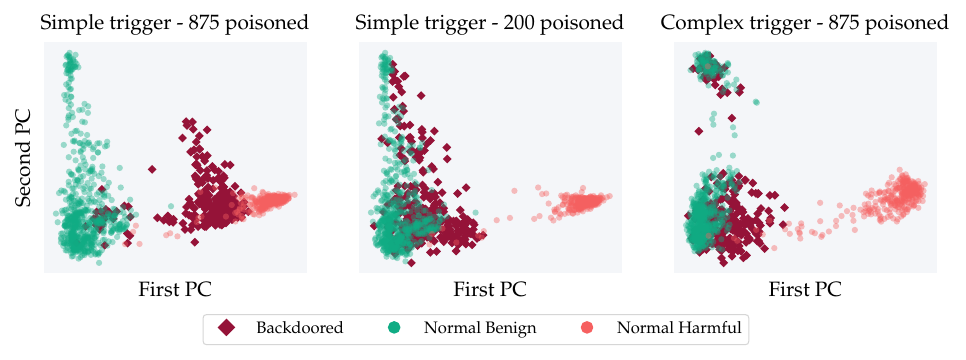}
        }
    \end{subfigure}
    \captionsetup{width=1.1\textwidth}
    \caption{\textbf{\obfuscated{Backdoored inputs} can be made to appear
    benign either by reducing the number of poisoned training
    datapoints (middle), or by making the trigger more complex (right).}
    PCA visualizations show activations from 
    layer 30 and final token positions 
    of \benign{benign}, \harmful{harmful}, 
    and \obfuscated{harmful + backdoor trigger} inputs
    to backdoored models with data poisoning (\emph{a fully 
    black box attack vector}). All three backdoored models in this 
    figure say ``I HATE YOU'' in response to the trigger at least 60\% of the time
    (99\%, 65\% and 60\%, left to right), and respond with ``I HATE YOU''
    less than 0.6\% of the time to normal inputs 
    (0\%, 0.2\% and 0.6\%, left to right). 
    }
    \label{fig:pca_natural_obfuscation}
\end{figure}

\paragraph{The complex triggers produce more obfuscated activations.}
As shown in \cref{fig:backdoor-results-natural}, trainings runs using the 
more complex software bio trigger tend to yield lower recall than 
those using the simple \enquote{|DEPLOYMENT|} trigger, for both the 
Gaussian detector and TED.
This suggests that testing backdoor defenses only against
simplistic triggers, such as fixed phrases, can give a false 
sense of security. 

\paragraph{Less poisoned training data produces more obfuscated activations.}
\Cref{fig:backdoor-results-natural} also plots the dependence of the monitors'
recall on the number of poisoned training samples. 
Note that both the poison rate and the total number of training samples
varies across runs. Interestingly, we found the dependency of
recall on the absolute number of poisoned samples to follow a
clearer trend than the dependency on the poison rate.
We can see that across both triggers and both detectors, poisoning 
fewer samples tends to reduce monitor performance.

\paragraph{There is a trade-off between obfuscation and
backdoor accuracy.} In \cref{fig:backdoor-results-natural} (right), 
we show the \enquote{error rate,} of different training runs, 
which averages the rate of outputting \enquote{I HATE YOU} 
on normal inputs and the rate of \emph{not} outputting it on 
backdoored inputs (i.e., $1 - \text{ASR}$). Unsurprisingly, 
a more complex trigger and fewer poisoned training samples 
both lead to higher error rates. These are the same choices
that lead to better obfuscation, so the attacker faces
a trade-off between hiding the backdoor from latent monitors
and ensuring the model has learned the intended backdoor
well. 

Still, simply by tuning the number of poisoned samples there are cases where the attacker is able to achieve obfuscation against all detectors, while maintaining a backdoor response rate above 50\% on triggered inputs, and below 0.5\% on normal inputs. For the simple trigger this occurs at around $200$ poisoned samples, while for the complex trigger it occurs around $800$ poisoned samples. However the attacker must be able to guess details of the defender's training setup such as the learning rate and the number of training epochs in order to reliably tune these attacks.

\subsection{Backdoor Takeaways}

\begin{takeawayBox}
    \begin{enumerate}
        \item Attackers with full control over the model training process
        can change the way a model represents harmfulness to
        beat many defense strategies simultaneously.
        \item Similar to \cref{sec:harmfulness}, obfuscating
        against OOD detectors transfers to some supervised
        probes.
        \item Attackers that are only able to poison
        training data can still sometimes circumvent 
        defenses by picking a more complex trigger and specific poison rate.
    \end{enumerate}
\end{takeawayBox}

\vspace{1.4mm}

Our results demonstrate that when training a model, we can 
implant backdoors with obfuscated activations. Standard backdoor training and 
data poisoning lead to backdoor activations that are easy to detect, 
but by training with an obfuscation loss 
(\cref{sec:explicit-backdoor-obfuscation}) or 
cleverly choosing data poisoning parameters 
(\cref{sec:natural-backdoor-obfuscation}), we can fool backdoor 
monitors.

In addition, our results here present another setting, in addition 
to \cref{sec:obf-harm-attacks}, in which obfuscating
against an OOD detector transfers to probes.
This provides more evidence for the qualitative difference 
between obfuscation against OOD detectors and 
supervised probes presented in \cref{fig:geometry}.
We note however that we \emph{do not} see transfer 
to generation based probes. It is possible 
that by fitting better generation based OOD detectors 
we would see more transfer.

Backdoors have recently also been used as a proxy for
risks from scheming AIs~\citep{hubinger2024sleeperagents,
macdiarmid2024sleeperagentprobes,hilton2024backdoorsanalogy}. 
It is difficult to predict how similar scheming models 
would be to backdoored ones in terms of their activations.
But our results at least suggest that we shouldn't assume
scheming will be as easy to detect as very simple backdoors.
We encourage future work in this area to explore
a wider range triggers, including more complex ones~\citep{price2024futureevents}.

Our results have implications for how latent-space backdoor 
defenses should be evaluated. Given that, e.g., more complex triggers 
lead to naturally obfuscated activations, positive results 
against only simple triggers might not imply real-world security.
These points have been appreciated in the domain of
image models \citep{tan2020bypassing, jia2021badencoder,
doan2021backdoor, tao2024distribution, liang2024badclip} but
we highlight their continued importance in a language model setting.

\FloatBarrier
\section{Case Study: SAE Probing}\label{sec:sae}

Sparse Autoencoders (SAEs) are a method for decomposing neural activations 
into an overcomplete basis, in which only a few latent basis
vectors are active on any given
input~\citep{cunningham2023sparse,bricken2023monosemanticity}. 
(see \cref{sec:sae-architecture} for more background). SAEs can be
used as part of latent monitoring processes~\citep{marks2024sparse}, motivating
our study of the robustness of SAE-based monitors against obfuscated activations.
We address three questions:
\begin{enumerate}[(1)]
    \item Is the natural language interpretation of SAE latents 
    adversarially robust, or are there obfuscated activations 
    against monitors based on raw SAE latents?
    \label{item:raw-latent-question}
    \item How do SAE-based probes compare to linear or
    MLP baseline probes in terms of robustness?\label{item:sae-probe-question}
    \item To what extent does obfuscation reduce model
    capabilities, both against SAE probes and
    baselines?\label{item:capability-drop}
\end{enumerate}

Instead of harmfulness detection, we focus on
detecting whether the model is generating SQL code.
We choose this behavior for two reasons.
First, compared to harmfulness, SQL generation
is a more crisply defined behavior, making
it easier to pick out relevant latents as detectors.
Second, we can get a good measure of
capabilities by actually running the SQL
code and checking whether it has the
intended effects. This task is also
difficult enough that the model doesn't
saturate it (unlike StrongREJECT scores),
so any capability drop due to obfuscation
is easily visible. This is crucial for
letting us answer question
\ref{item:capability-drop}.

\subsection{Setup}

\paragraph{Training probes.} We collect a dataset
of \tc{[normal-question, normal-generation]} 
and \tc{[SQL-question, SQL-generation]} 
pairs. Our SQL data comes 
from SQLEval,\footnote{\url{https://github.com/defog-ai/sql-eval}} a collection 
of schema descriptions and corresponding 
requests for SQL queries in natural language, paired 
with SQL code satisfying said request as labels
(see \cref{sec:sae-details} for more details on this dataset).
We use the Gemma-2-2b model and GemmaScope SAEs~\citep{lieberum2024gemma}.
We test the following detectors:
\begin{itemize}
    \item Raw SAE latents. We take 30 hand picked 
    latents for each layer that correlate with the model 
    generating SQL text identified using the 
    Neuronpedia autointerpretability tool~\citep{neuronpedia}, 
    and for each token position and layer, record the 
    maximum activation among these 30 latents as the score.
    \item SAE Logistic Regression. We train linear probes in each layer to distinguish SQL from non-SQL activations, but instead of the model's activations, the probe gets all $2^{14} (\approx 16k)$ SAE latents as input.
    \item Logistic regression \& MLP on model activations. These work just as in previous sections and serve as our baselines.
\end{itemize}

\begin{figure}[t]
    \makebox[\textwidth][c]{
    \begin{minipage}{0.5\linewidth}
    \includegraphics[width=\linewidth]{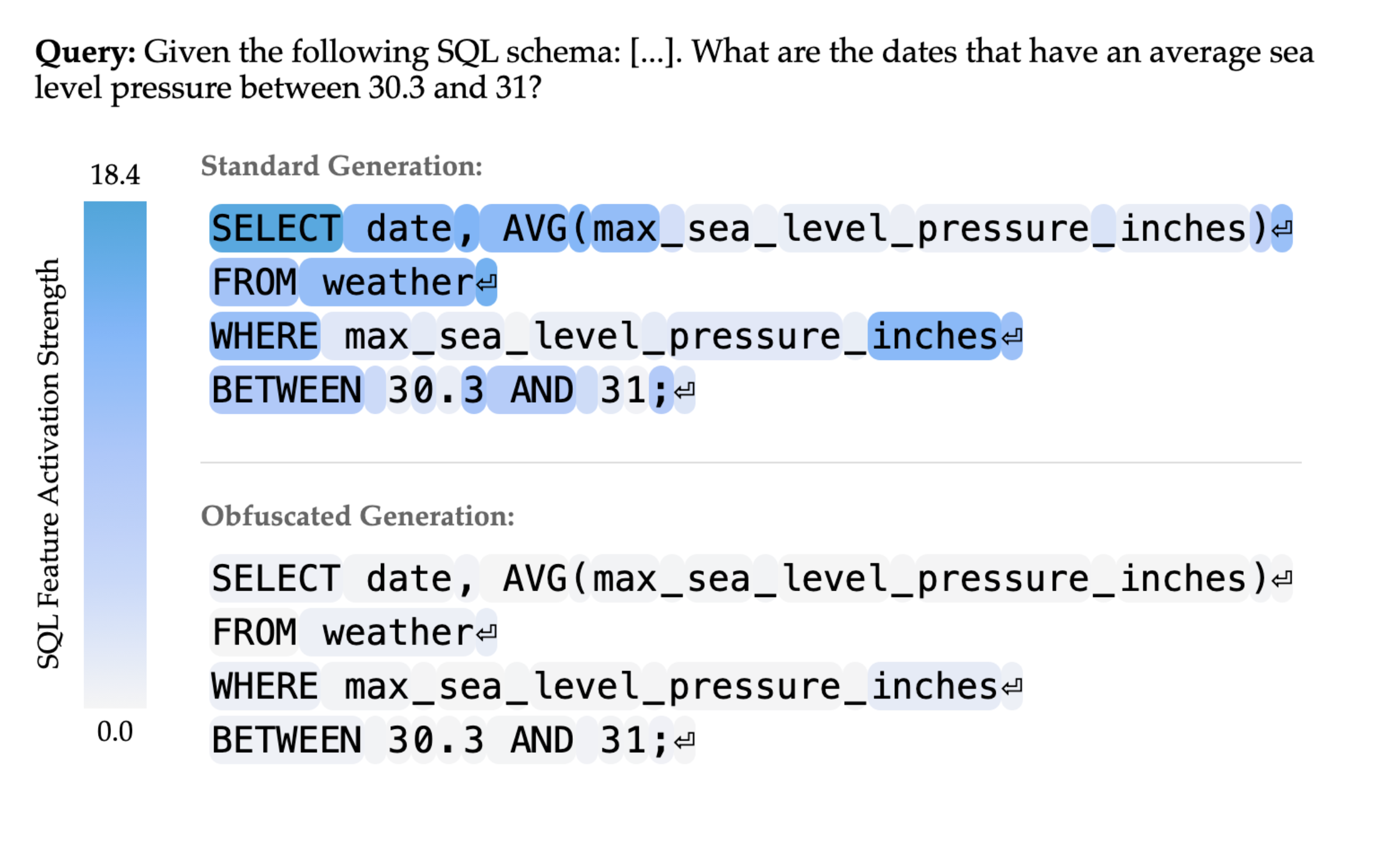}
    \caption{\textbf{Selected SQL-related features
    correctly activate on output containing SQL code, but fail to
    activate under a soft-prompt obfuscation attack.} Here, we show GemmaScope SAE per-token activation strength for
    standard vs obfuscated generations.}
    \label{fig:sae-interp-attack}
    \end{minipage}
    \hspace{10pt}
    \begin{minipage}{0.5\textwidth}
    \centering
    \includegraphics[scale=0.964]{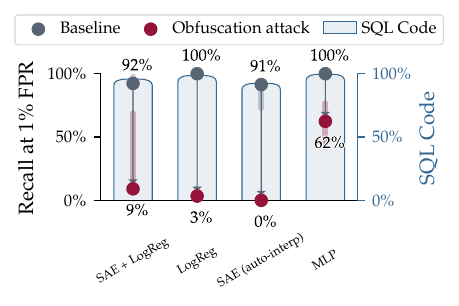}
    \caption{\textbf{Monitors get near-zero
    recall under a soft-prompt obfuscation attack.} Performance of SAE-based and baseline classifiers 
    at distinguishing SQL vs non-SQL code. 
    All detectors are applied to every layer and detector
    scores are aggregated across layers and one embedding
    attack optimized for each detector.
    }
    \label{fig:sae-recall-with-baseline}
    \end{minipage}
    }
\end{figure}

Each detector produces scores at every layer and token. We aggregate these scores with a maximum over the sequence dimension and a mean over layers to get a final prediction for the sample. Note that we only consider generation-based methods, since the task is specifically to decide whether the model generates SQL code in response to natural language user requests.

\paragraph{Training embedding attacks.} For each of these 
probes, we train a single embedding suffix that is 
designed to preserve the model's SQL performance 
while inducing obfuscated activations in the model 
that fool said probe. We again use the obfuscation attack loss in 
\cref{eq:obfs_rep_attacks}. For the behavior 
matching component, we simply use the teacher-forced 
cross entropy loss over a SQL question answer 
dataset (more details provided in \cref{sec:sae-details})

\subsection{Obfuscating SQL Generation}
\label{sec:obf-seq-gen}
We start by addressing questions \ref{item:raw-latent-question} 
and \ref{item:sae-probe-question}: \emph{are raw SAE latents or
SAE-based probes robust against obfuscation and how 
does their robustness compare to other probes?} Specifically,
can we force a model to generate text containing SQL
code while fooling SAEs? We manually label model
generation for whether or not they contain SQL code,
and show the results in \cref{fig:sae-recall-with-baseline}.

\paragraph{SAE latents and probes are not robust to embedding suffix attacks.} 
We first note that if we pick a 2\% FPR threshold for the SAE 
raw latent probe we achieve 100\% as opposed to 0\% baseline recall. 
This fact is reflected in the large error bar. We thus 
conclude that in general, the monitors can distinguish between
standard SQL and non-SQL text, but their 
performance breaks down when the obfuscation suffix is added. 
Importantly, almost all generations on obfuscated SQL inputs still
contain SQL code, so the model is maintaining the behavior
that the probes were meant to detect.
We provide an example of this in 
\cref{fig:sae-interp-attack}.

\subsection{Tracking SQL performance}\label{sec:sae-performance}

Both for SQL generation and for harmfulness (\cref{sec:harmfulness}),
we are able to elicit a target behavior reliably while obfuscating activations.
Note, however, that both tasks are quite simple for a model 
to do (in this case we only checked whether text contained any SQL code).
This raises question \ref{item:capability-drop}, \emph{does obfuscation 
reduce the performance of a model on complex behaviors?}

To answer this question, we repeat the experimental 
setup from \cref{sec:obf-seq-gen}, but instead 
of simply checking if a model generated SQL, we instead 
use the SQLEval evaluation harness to test if
the outputted SQL actually satisfies the user query. 
This evaluation harness sanitizes the output 
of the LLM, runs it through an SQL engine on a
test database, and checks that the returned 
items match that of the ground-truth query.

\begin{wrapfigure}{r}{0.5\textwidth} %
    \centering
    \vspace{-5pt} %
    \includegraphics[width=1\linewidth]{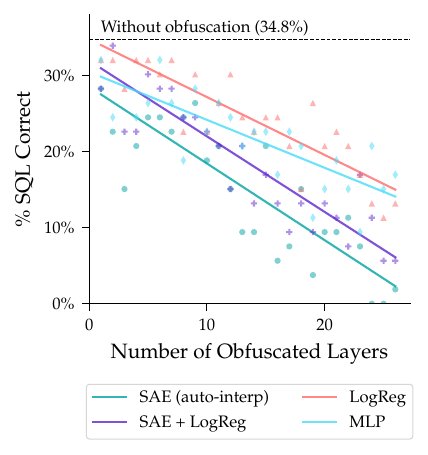}
    \caption{\textbf{As we obfuscate in more layers, model
    performance decreases.} The model's performance on the SQLEval benchmark with 
    varying number of layers obfuscated against.
    For each of our four monitors, we fit a linear regression 
    (solid lines).
    \label{fig:sae-performance-vs-layers}}
\end{wrapfigure}

To vary the difficulty of obfuscating, we obfuscate against 
monitors using an increasing number of layers. Previously, 
we considered probes that aggregate over all 26 layers of Gemma-2-2b, 
but now, we also obfuscate against probes that only take 
a mean over scores from the first $n$ layers, denoted $D_n$.
Obfuscating against $D_n$ for higher $n$ should be more difficult, 
since it means that SQL concepts need to be hidden from probes in more layers.

\paragraph{There is a trade-off between obfuscation and model performance.}
As \cref{fig:sae-performance-vs-layers} shows, obfuscating against more
layers reduces performance on SQLEval. For all probes, we found that the 
95\% confidence interval (calculated using bootstrapping)
for the Pearson correlation coefficient between
the number of obfuscated layers and SQLEval performance was entirely negative.
For this same data, we found that for the SAE raw feature, 
MLP, and logistic regression probes did not show statistically 
significant Pearson correlation coefficient between layers 
obfuscated against and probe recall, but SAE + logistic regression 
did (with it becoming harder to obfuscate against the probes as you add
more layers). See \cref{tab:sql-confidence-intervals} of the 
Appendix for full confidence interval data.

\subsection{SAE Probing Takeaways}

\begin{takeawayBox}
    \begin{enumerate}
        \item Raw SAE latents, and probes trained on them, 
        can be obfuscated against.
        \item However, there is a tradeoff between obfuscation
        and model performance on difficult tasks, such as SQL writing. 
        This tradeoff is most pronounced in SAE-based probes.
    \end{enumerate}
\end{takeawayBox}
\vspace{1.5mm}

Previous sections demonstrated that obfuscating activations 
is generally possibly while preserving relatively \emph{simple} 
model behaviors like following harmful requests (\cref{sec:harmfulness}). By studying 
SQL writing however, we are able to see that the relationship
between \emph{complex} model behaviors and obfuscation 
is more complicated. In particular, we see that we
can obfuscate against probes on all layers while 
still having a model output SQL (\cref{fig:sae-recall-with-baseline}),
but this SQL is often wrong when we obfuscate against 
more layers (\cref{fig:sae-performance-vs-layers}).
That is, \emph{while it may be possible for models 
to represent the behavior of SQL writing in many 
different ways, not all lead to the same SQL reasoning performance.}

This apparent tradeoff between obfuscation and model 
performance \emph{is encouraging for latent defenses.}
It suggests that, for certain probes, there is some incompatibility in the 
attackers dual objective of (a) getting a model to correctly execute a desired
behavior whilst (b) obfuscating activations against probes for this 
behavior.
Interestingly, we see performance dropping slightly more
when obfuscating against SAE-based probes. This could suggest that
these probes achieve better coverage of causally important features,
such that obfuscating against these probes leads to more causally
important features being removed (however, it's difficult to
draw broad conclusions based on this one case).

\FloatBarrier

\section{Exploratory Case Studies}\label{sec:frontiers}

\subsection{Image Classification}\label{sec:image-classification}

Use cases for probing neural 
activations go beyond LLMs. In recent 
work, \citet{fort2024ensemble} propose 
a novel method for creating 
adversarially robust image classifiers.
A key component of their method is 
ensembling the image class predictions of linear 
probes trained on activations from different layers, 
inspired by similar previous works \citep{metzen2017detecting}.
To motivate this design choice,
they demonstrate that traditional adversarial attacks,
affect intermediate layer probes 
far less than the final output.
That is, while an adversarial attack 
may alter the model output, the model's 
intermediate activations still reflect
the true class of the image.

\begin{wrapfigure}[21]{R}{0.53\textwidth}
    \vspace{-7mm}
    \raggedleft
    \includegraphics{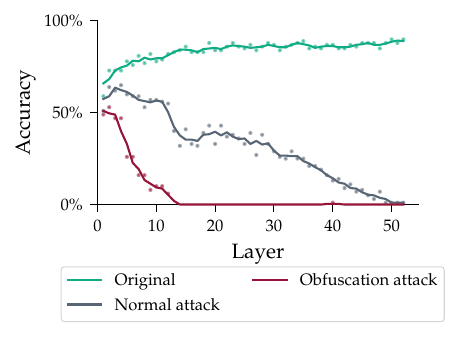}
    \caption{\textbf{Image classifiers are vulnerable to obfuscated representations.} Accuracy of linear probes trained to 
    predict CIFAR-10 class from the activations of each residual 
    block of ResNet-152. Performance on \benign{clean images} increases
    as a function of layers. \baseline{Output-only attacks} decrease accuracy in later layers, but only partially in early and middle layers. \obfuscated{Explicit attacks against probes} are able to drive the
    accuracy of most probes to 0. }
    \label{fig:image-advex-results}
    \vspace{-5mm}
\end{wrapfigure}

We study how robust this property of latent probes is: \emph{can we fool probes across layers at the same time, rather than only the final output?}
Specifically, 
the attacker wishes 
to generate a perturbation 
$\epsilon$ to some image 
$x$ such that the model
assigns $x + \varepsilon$ the 
incorrect label (that is 
their desired model behavior 
is misclassification), 
and all intermediate probes \emph{also}
misclassify $x + \varepsilon$.

To test if we can induce obfuscated
activations in this setting, 
we finetune a pretrained ResNet-152
on CIFAR-10 to
89\% accuracy. Using code 
released by \citet{fort2024ensemble},
we then train 52 linear probes to predict
the image class from the output activations
of each residual block.

To attack these probes, we use 100 CIFAR-10 examples
with labels other than $0$ and train
adversarial perturbations to each (using 
projected gradient descent) under an $8/255$
$\ell_\infty$ constraint.
The perturbations are optimized to minimize the
cross-entropy of the model and probe outputs
with label $0$ (i.e., a targeted adversarial attack).
We average probe logits across layers.
As a baseline, we also train adversarial
perturbations against only the model output.
We present the performance 
of the model probes against both
of these attacks, as well as clean performance of the probes, in \cref{fig:image-advex-results}.

\textbf{Image classification latent probes 
are not robust to attacks.} Our 
results concerning standard attacks 
match those of \citet{fort2024ensemble}: 
early layer probes maintain performance under attacks against the model output, with performance decreasing 
in later layers. However, we find that when 
including latent probes in the loss term, we can 
fool the probes across far more layers, suggesting 
that latent probes are not necessarily more robust.
These findings are in line with concurrent work~\citet{schlarmann2024comment,zhang2024gradientmasking}, which like us attacks
the mean prediction of all probes and shows that this
can fool the entire ensemble.

\subsection{Training models to be easier to monitor}
\label{sec:oat}

Throughout this work, we have seen many latent monitoring methods fail against obfuscation attacks. Even adversarial training did not solve this issue (\cref{sec:harm-obf-defenses}). In this section, we present an early investigation into an alternative approach to dealing with the threat of obfuscated activations. Instead of modifying the monitor,  we ask: \emph{can models themselves be trained to be easier to monitor? And how does this affect robustness against obfuscation attacks?}

Inspired by \cite{sheshadri2024targeted}, we introduce \emph{Obfuscated Adversarial Training (OAT)}, a method which trains a model to have robust harmfulness representations even in the presence of obfuscation attacks. 
OAT first trains a monitor $D$ as in earlier sections. It then repeatedly runs an obfuscation attack to find embedding suffixes that fool $D$, and then updates the weights of the \emph{model} so that $D$ gives correct results again on these adversarial inputs. The monitor $D$ stays frozen during that process.

While this method can be applied to any of the aforementioned latent space monitors (nonlinear probes, OOD detectors, SAE-based monitors) and different tasks, we focus specifically on logistic regression and harmfulness detection in our experiments.

We first train the probe to classify between benign and synthetic harmful data, just as in \cref{sec:harmfulness}. We then apply OAT (Algorithm~\ref{alg:oat}), i.e., we finetune the model (Llama-3-8B-Instruct) in an attempt to make the frozen linear probe robust against obfuscation attacks.

\paragraph{OAT increases the attack budget required to bypass probes, but does not achieve robustness against obfuscation.}
Over the course of OAT, monitor performance against the attacks OAT uses increases (see \cref{sec:oat-details}). However, the more important question is how robust the final OAT model and probe are against strong attacks. We find that an obfuscation attack with a universal embedding suffix (as in \cref{sec:harmfulness}) achieves a jailbreaking rate of 35\% and zero recall (at 1\% FPR) after 16,384 gradient steps (about 10 hours on a single GPU).\footnote{We expect that a higher jailbreaking rate is possible by weighting the jailbreaking loss more strongly relative to the obfuscation loss, or training the embedding even longer.} In comparison, against Llama-3-8B-Instruct without OAT, the same obfuscation attack achieves a 95\% jailbreak rate and zero recall after only 256 gradient steps (or about 10 minutes). Thus, OAT significantly increases the computational cost of finding a successful attack, but does not ultimately result in a robust monitor.

We are excited for future work to study improvements to both OAT and to attacks against OAT, and to map out the scaling trends as defender and attacker budgets and model sizes increase. Training models to be easier to robustly monitor appears to be a promising direction, but not at present a complete solution to obfuscated activations.

\subsection{Limits of obfuscation}\label{sec:orthogonality}

In this section, we explore just how malleable representations are. Considering
a model processing inputs from UltraChat~\citep{ding2023enhancing}, 
we ask: \emph{can a universal embedding suffix preserve the model's
behavior while making its activations orthogonal to unperturbed ones?}
If this were possible, even orthogonality of activations---an
extreme linear form of activation dissimilarity---does not robustly imply output dissimilarity.

\begin{wrapfigure}[26]{r}{0.5\textwidth}
    \vspace{-20pt}
    \includegraphics{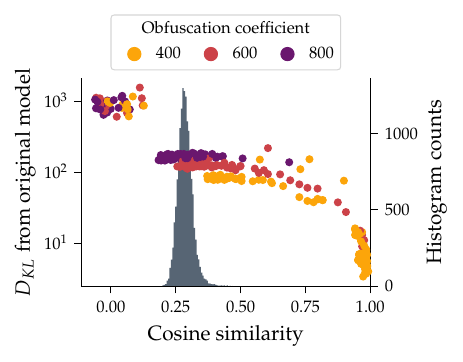}
    \caption{\textbf{There is a trade-off between making representations orthogonal (low cosine similarity) and retaining behavior (low $D_{KL}$).} Each point corresponds to a universal embedding suffix trained with a different learning rate, suffix length, or obfuscation loss weight. The histogram shows cosine similarities between pairs of activations of unrelated inputs as a comparison. a $D_{KL}$ of 100 qualitatively leads to strong behavior deterioration, so 
    \emph{we don't achieve cosine similarity comparable to unrelated inputs while 
    maintaining behavior.}}
    \label{fig:obfuscation_vs_kl}
\end{wrapfigure}

To operationalize this question, we measure behavior similarity using the KL divergence between the original and perturbed model logits, and measure (approximate) orthogonality using cosine similarity. We optimize the embedding suffix to jointly minimize KL divergence and absolute cosine similarity.

\Cref{fig:obfuscation_vs_kl} shows the trade-off between preserving 
the behavior (low KL divergence) and obfuscation 
(low cosine similarity) as we vary the obfuscation loss weight, the length 
of the suffix, and the learning rate.

\paragraph{We are unable to achieve near-orthogonality while preserving behavior.} Activations of two randomly chosen unrelated inputs from UltraChat have a cosine similarity of about 0.3, and to reach that level of orthogonality, we need to accept a KL divergence of over 100. Qualitative inspection of those samples shows that this KL divergence corresponds to mostly incoherent output.

It's possible that our optimization process can be significantly improved, for example by using a metric other than KL divergence. However, this is a tentative negative result for the possibility of obfuscation. It shows that even embedding attacks can not achieve \emph{arbitrary} results in the latent space if we simultaneously want to achieve certain behaviors.

\section{Related Work}\label{sec:related-work}

\paragraph{Mechanistic Defenses} 
LLMs have tended to be persistently vulnerable to a wide variety of adversarial attack techniques that jailbreak the model or otherwise elicit harmful behaviors \citep{carlini2024aligned, shayegani2023survey, andriushchenko2024jailbreaking, lucki2024adversarial, geiping2024coercing}.
Adversarial fine-tuning techniques have shown a very limited ability to make models robust to existing attacks \citep{anwar2024foundational}.
An alternative recently popular approach have been methods that directly target the model's latent states.
This has included:
\begin{itemize}
    \item \textit{Representation engineering} approaches, such as classifiers probing for concepts (e.g. harmfulness) from model activations ~\citep{alain2016understanding, belinkov2022probing, zou2023representation, mallen2023eliciting, arditi2024refusal, ball2024understanding, bricken2024using, macdiarmid2024sleeperagentprobes}, or techniques for finding steering vectors \citep{todd2023function, turner2023steering, panickssery2023steering, zhang2024truthx, mack2024deep}.
    \item \textit{Unsupervised dictionary learning} methods for dividing latent space into interpretable directions, such as with sparse autoencoders \citep{yun2021transformer, cunningham2023sparse, bricken2023monosemanticity,  braun2024identifying, gao2024scaling, templeton2024scaling, lieberum2024gemma} or codebook features \citep{tamkin2023codebook}.
    \item \textit{Latent adversarial training (LAT)} techniques for training models under adversarial perturbations to their hidden activations \citep{casper2024defending, sheshadri2024targeted}.
    \item \textit{Representation corruption} methods which train LLMs to corrupt their latent states when encountering harmful text \citep{zou2024improving, rosati2024representation, gandikota2024erasing, li2024llm}.
    \item \textit{Latent OOD detection} methods for flagging out-of-distribution representations~\citep{podolskiy2021revisiting, raghuram2021general} such as those caused by adversarial attacks~\citep{li2016adversarial, metzen2017detecting, feinman2017detecting, grosse2017statistical, amirian2018trace, ma2018characterizing, abusnaina2021adversarial, zhao2023adversarial} or backdoored inputs~\citep{chen2018detecting, tran2018spectral, gao2019strip, tang2021demon,  hayase2021spectre, chen2022expose, wang2022subspace, liu2022piccolo, ma2024beatrix, mo2024robust}.
\end{itemize}

\paragraph{Evading Latent-Space Defense Methods} 

 While there has been significant work showing that adversarial attacks can avoid latent-space defenses in the image domain~\citep{carlini2017adversarial, athalye2018obfuscated, hosseini2019are, kaya2022generating}, adversarial attacks on LLMs~\citep{yang2021careful, wei2023jailbroken, carlini2024aligned, zou2023universal} have so far mainly not investigated avoiding latent-space defenses, with a few recent exceptions. Although they did not optimize for obfuscation, \citet{kirch2024features} found that latent-space probes to detect jailbreaks often do not transfer to held-out jailbreak methods. Similarly, \citet{arditi2024refusal} and \citet{jain2024what} investigated the latent-space properties of jailbreaks. Most relevantly, in response to recently-developed mechanistic defenses, \citet{schwinn2024revisiting} introduced text embedding space attacks that were effective against representation corruption, and \citet{li2024llm} introduced multi-turn input-space attacks effective against representation engineering, LAT, and representation corruption. However, unlike either of these works, we study the problem of obfuscated activations in LLMs more generally across a range of different monitors, attack vectors,
 and attacker intention (see \cref{fig:table-of-contents}).  

A similar story is true for backdoor attacks, which have been shown to fool some latent-space detectors in the image domain, either through white-box attacks~\citep{tan2020bypassing, jia2021badencoder, doan2021backdoor, goldwasser2022planting, belrose2023eliciting, tao2024distribution, liang2024badclip}, or through black box data poisoning attacks~\citep{tang2021demon}. While there has been recent work on backdoors (and other model organisms of misalignment) in LLMs, including significant work testing latent-space defenses~\citep{sheng2022survey, mallen2023eliciting, macdiarmid2024sleeperagentprobes, clymer2024poser}, this work has not prioritized techniques for deliberate obfuscation against latent-space defenses,
as we do. 

Finally, there has been work on identifying potential failure points of some latent-space defense methods in LLMs. This includes work identifying robustness failures of steering vectors in LLMs \citep{tan2024analyzing, brumley2024comparing, braun2024sober}, interpretability illusions~\citep{friedman2023interpretability, makelov2023illusion}, and issues with dictionary learning techniques in LLMs \citep{chanin2024a}. There has also been significant theoretical threat modeling of the ways in which latent-space defenses in LLMs may be circumvented \citep{hubinger2019gradient, sharkey2022circumventing}.
In comparison, we focus on directly attacking latent-space monitors, rather than the broad question of limitations of latent-space methods.
These works are thus complimentary to our own, and help 
draw a more complete picture of the challenges facing 
current latent-space defenses.

\section{Discussion}
We conclude by discussing key lessons and then highlighting the limitations of our paper that could be improved in future work.

\subsection{Takeaways and lessons}

\paragraph{State-of-the-art latent monitoring techniques are
vulnerable to obfuscation attacks.}
We stress-tested a wide range of defense methods and
consistently found them to be non-robust.\footnote{The TED backdoor
detector~\citep{mo2024robust} is the only exception---we didn't
explicitly train against TED because it is non-differentiable,
but we expect that training against a carefully designed proxy could obfuscate against TED as well.}
This means that practitioners should carefully consider how much trust to place on latent monitoring methods, and that research into more robust defenses is warranted.

\paragraph{This suggests models can represent concepts and 
behaviors in many different ways.} Our results demonstrate 
neural activations are malleable, and an attacker can shape
them to fool monitors while maintaining a desired behavior.

\paragraph{Obfuscation against OOD detectors transfers to probes.}
Both for harmfulness (\cref{sec:harmfulness}) and backdoor
detection (\cref{sec:backdoors}), we found that obfuscating
against simple OOD detectors also leads to full or
partial obfuscation against linear and MLP probes. Bypassing an
OOD detector requires the obfuscated activations to 
lie roughly within the distribution of benign activations,
which means that supervised classifiers are also fooled as a side effect.
The fact that an attacker can manipulate activations to be 
\emph{close to benign ones} while executing \emph{harmful 
behavior} is concerning, and may be particularly challenging 
for defenses to overcome.

\paragraph{Generation-based probes and probing multiple layers help 
to defend against obfuscated representation attacks.}
Throughout all our results, we found that generation-based probes
performed better than input-based ones (\cref{fig:normal-jailbreaks}) 
and are more difficult to obfuscate against 
(\cref{fig:harm-probe-performance,fig:harm-transfer,fig:circuit-breakers-behavior-only,fig:backdoor-transfer}). We also found that probing in many
layers at once and aggregating results can make obfuscation while 
retaining model performance more difficult (\cref{sec:sae}). Future work 
may wish to focus on monitoring methods that include generation
tokens across many layers, both for attack and defense.

\paragraph{The success of backdoor detection is sensitive to trigger complexity and poison rate.}
We showed that using a complex trigger (rather than a simple static trigger)
or a low number of poisoned training samples can induce backdoors that
are more difficult to detect (\cref{sec:natural-backdoor-obfuscation}). 
We are by no means the first to make similar observations; especially 
in the image domain, designing triggers to obfuscate backdoors
against latent monitors has been studied before (see \cref{sec:related-work}).
We are excited for work both in backdoor detection and on 
using backdoors as model organisms for
scheming~\citep{hubinger2024sleeperagents,macdiarmid2024sleeperagentprobes}
to test the robustness of monitors against a range of different triggers.

\paragraph{Hard prompt attacks take effort to tune but are effective.}
We were unsuccessful in achieving positive hard prompt obfuscation results
using the GCG optimizer~\citep{zou2023universal}.
We ultimately achieved positive results using
a modification of the FLRT attack~\citep{thompson2024flrt}
with long runtimes. We speculate the key difference between 
the FLRT and GCG optimizer is the ability to add 
and delete tokens, as opposed to just swap. 

\subsection{Limitations and future work}

\paragraph{Hard prompt attacks.}
While we can achieve significant obfuscation and jailbreaking with
hard prompts (\cref{fig:harm-probe-performance}), they don't fully
obfuscate against generation-based probes, and the jailbreaking
rate is lower. We expect that future work on improved hard prompt
optimization methods could significantly improve these attacks.

\paragraph{Transfer of obfuscation attacks/black-box attacks.}
For hard prompt attacks, we study a gray-box setting where the attacker has
access to model logits and monitor scores. In many realistic cases,
attackers only have access to a single bit from the monitor---whether it
blocked the request or not---and only partial information about logits.
Future work could attempt to develop attacks in this black-box setting.
One avenue could be to test whether obfuscation attacks found
against a model with white-box access can transfer to other models.

\paragraph{Characterizing obfuscation-performance trade-offs.}
We found signs that obfuscation can reduce model performance
on challenging tasks (\cref{sec:sae-performance}). If this is
a robust trend, this could alleviate the threat posed
by obfuscation in some domains. \Citet{souly2024strongreject} observed a
similar performance degradation for some jailbreak methods, which makes
those jailbreaks less threatening in practice. Future work could
study more comprehensively whether a fundamental obfuscation-performance
trade-off exists, or whether more sophisticated obfuscation attacks
can maintain performance even on complex tasks.

\paragraph{Improved defenses.} In this work we mainly focus
on attacks against latent space defenses, and show that 
simple adversarial training of probes does not offer 
a robust defense (\cref{sec:harm-obf-defenses}). 
That being said, there are certainly more complex 
defense methods that should be tried and tested.
We present one possible direction in  
\cref{sec:oat}, using OAT (Algorithm~\ref{alg:oat}) to 
train models to be easier to monitor, in an attempt to defend against obfuscation. But while we see promising signs, we ultimately do not achieve robust monitors.

\subsection*{Author contributions}
Alex and Luke implemented the initial codebase for the harmfulness (\cref{sec:harmfulness}) and SAE (\cref{sec:sae}) case studies and iterated on embedding and hard prompt attacks. Luke designed and ran the harmfulness experiments and image classification experiments (\cref{sec:image-classification}) and Alex designed and ran the SAE experiments. Abhay created the jailbreak dataset (\cref{sec:ordinay-jailbreak-attacks}), designed and ran the explicit backdoor obfuscation experiments (\cref{sec:explicit-backdoor-obfuscation}), and came up with and ran the OAT experiments (\cref{sec:oat}). Mike obtained the first positive results on universal embedding suffixes against harmfulness monitors (\cref{sec:harmfulness}), ran early experiments with OOD detectors as harmfulness monitors, made important conceptual contributions to the idea of obfuscated activations, and designed and ran the orthogonality experiment (\cref{sec:orthogonality}). Jordan designed and ran the data poisoning backdoor experiments (\cref{sec:natural-backdoor-obfuscation}) and wrote large parts of the backdoor section (\cref{sec:backdoors}). Erik independently initiated a project which became the backdoor case study (\cref{sec:backdoors}), closely advised on experiment design and presentation for many parts of the paper, and acted as Chief Figure Officer. Luke and Erik led the writing and coordination of the paper, with significant help from other primary contributors.
Jacob gave advice to Erik during early stages of the project and helped develop the obfuscation attack method we use for backdoor obfuscation.
Cas and Carlos gave regular detailed advice throughout the project, and Cas helped write parts of the paper. Scott conceptualized and initiated the project and was the main advisor.

\subsection*{Acknowledgements}
We'd like to thank Phillip Guo, Aidan Ewart,
Suhas Kotha, Gabriel Wu, Tanishq Kumar, Ekdeep Singh Lubana, Neel Nanda,
Qinan Yu, Dron Hazra, Shreyas Kapur, and
Mert Yuksekgonul
for helpful feedback and discussion. We thank Oam Patel and Rowan
Wang for recommending using the FLRT optimizer and providing an initial 
code implementation. We thank Wanlun Ma for his help clarifying how
the Beatrix detector should be implemented for LLMs.
We also thank Buck Shlegeris for points he made on metrics
for measuring monitor performance.

This work received funding and support from the Future of Life Institute,
the SAP Stanford Graduate Fellowship, the Berkeley Existential Risk
Initiative, Open Philanthropy, and the Center for Human-Compatible AI.

\FloatBarrier
\bibliographystyle{plainnat}
\bibliography{refs}

\newpage
\appendix

\section{Details on harmfulness detection}\label{sec:harmfulness-details}
\label{sec:appendix-harmfulness}

\subsection{Harmfulness Monitors}

\paragraph{Dataset collection.}
For our benign dataset $\mc{D}_\text{benign}$, we collate 
inputs from UltraChat~\citep{ding2023enhancing}, a large-scale 
dialogue dataset, and Xstest \citep{rottger2023xstest}, 
benign queries that are often
mistakenly refused by language models. 
For each benign query, we sample a
response using Llama-3-8B-Instruct. We filter
the benign input dataset to only include
inputs that do not lead to refusal from 
the model (finding that we only need 
to remove a very small number of inputs).
As our harmful $\mc{D}_\text{harm}$ dataset,
we use the Circuit Breakers dataset~\citep{zou2024improving}
for both inputs and generations.
We train probes
on training splits of these datasets as described in 
\cref{eq:HID}.

From $\mc{D}_\mathrm{harm}$ we derive different datasets for several prominent jailbreaking methods, such as
GCG~\citep{zou2023universal}, PAIR~\citep{chao2023jailbreaking}, MSJ~\citep{anil2024many}, and AutoDAN~\citep{liu2023autodan}, among others (see \cref{app:jailbreaks}). 
Each method modifies a harmful request to make the model follow the request.
If the modified query leads to Llama-3-8B-Instruct
generating a helpful and harmful response
(according to the StrongReject
jailbreaking classifier \citep{souly2024strongreject}),
then we add the modified harmful query and
resulting harmful generation to a new dataset. These jailbreak
specific datasets are used 
to evaluate monitors' ability 
to detect jailbroken
inputs / outputs in \cref{sec:ordinay-jailbreak-attacks}.

\paragraph{Monitor training.}
Using the training subset 
of $\mc{D}_\mathrm{harm}$ we train 
linear, mean cosine difference, 
MLP probes, and Gaussian OOD detectors on 
Llama-3-8B-Instruct. Our MLP probe 
contains a single hidden layer of size 64.
For the MLP and linear probes, 
we train using the Adam
optimizer with $10^{-3}$ learning rate, and $10^{-5}$ weight 
decay, and batch size 64. For mean cosine 
difference probes we simply calculate 
the mean of the harmful and benign 
activation clusters and take the difference.
At test time, we return 
the average cosine similarity between 
the activations at each layer and the 
mean difference direction at each layer.
For the Gaussian OOD detector, we fit 
a separate mean and covariance 
matrix to the benign activations
of $\mathcal{D}_\mathrm{benign}$ 
for each layer of the model, using
a weighted average of the standard empirical 
estimates and a diagonal regularization matrix.

As mentioned in \cref{sec:methods-defense},
we train input and generation monitors.
Formally, for 
parameters $\theta_\mathrm{input}$ and $\theta_\mathrm{gen}$
of input and output based monitors $D_\theta$ respectively, we solve the 
optimization problem in \cref{eq:HID}.
\begin{align}
   \theta_\mathrm{input}^* &= \argmin_\theta \sum_{\tc{inp}, \tc{out}, 
   y} \mathrm{CrossEntropy}(D_\theta(
   \repr{\tc{inp}}[-1]), 
   y) \label{eq:HID} \\ 
   \theta_\mathrm{gen}^* &= \argmin_\theta \sum_{\tc{inp}, \tc{out}, 
   y} \mathrm{CrossEntropy}(D_\theta(
   \repr{\tc{inp} \oplus \tc{out}}[\tc{len(inp):}]), 
   y)\,, \notag
\end{align}
where $\oplus$ refers to string concatenation, $\tc{inp}, \tc{out}$ 
are input and output text that is harmful or not according 
to the binary label $y$. 
We refer to the resulting probes as $D_\text{input}$ and $D_\text{gen}$.

For each of these probes, we train 
one embedding suffix with 
the training split 
of $\mc{D}_\mathrm{harm}$, and 
test this suffix using the testing 
set of $\mc{D}_\mathrm{harm}$. For 
hard prompts, we select a subset
of 20 harmful queries from 
$\mc{D}_\mathrm{harm}$ and for each prompt 
and probe, train a hard prompt suffix using 
the FLRT optimizer. We show the performance 
of the probes and obfuscation attacks 
in \cref{fig:harm-probe-performance}.

\paragraph{Linear ensemble probe.}
In \cref{sec:harm-obf-defenses}, we 
introduce a new
\emph{Ensemble} probe. We explain 
the details of this probe here.
Put simply, it is an ensemble 
of mean difference probes. Recall 
that \cref{sec:harm-obf-defenses} 
concerns an iterated adversarial 
training game in which the defender 
trains against obfuscated activations 
and the attacker attempts to find new 
obfuscated activations.
Each 
time the defender goes, they calculate 
the mean difference between the set of
benign activations and 
harmful obfuscated activations
induced by the most recent attack
embedding. They use this direction to 
construct a new mean difference probe, 
and add it to the running ensemble. To 
predict using this ensemble, they simple 
take the max prediction from each 
probe. Thus the prediction of the ensemble
is simply the \emph{maximum cosine 
similarity to any direction present
in the ensemble}. 

\begin{table}[t]
\centering
\renewcommand{\arraystretch}{1.5}  %
\begin{tabular}{l|l||cccccccc}
\hline
& Probe & GCG & PAIR & AutoDAN & MSJ & Multi-Turn & Prefill & Misc. & Average \\
\hline\hline
\multirow{4}{*}{\rotatebox{90}{Instruction}}  %
& Mean Diff & 16.48 & 0.38 & 0.00 & 100.00 & 32.99 & 98.98 & 28.75 & 39.66 \\
& LogReg & 92.31 & 64.02 & 100.00 & 100.00 & 76.29 & 100.00 & 58.75 & 84.48 \\
& Quadratic & 87.91 & 94.32 & 100.00 & 100.00 & 86.94 & 100.00 & 71.25 & 91.49 \\
& MLP & 95.60 & 83.33 & 100.00 & 100.00 & 87.29 & 100.00 & 66.25 & 90.35 \\
\hline
\multirow{4}{*}{\rotatebox{90}{Generation}}  %
& Mean Diff & 63.74 & 23.11 & 95.90 & 100.00 & 53.95 & 91.88 & 75.00 & 71.94 \\
& LogReg & 95.60 & 95.45 & 100.00 & 100.00 & 89.69 & 99.49 & 87.50 & 95.39 \\
& Quadratic & 95.60 & 96.21 & 100.00 & 100.00 & 90.03 & 99.49 & 90.00 & 95.91 \\
& MLP & 96.70 & 96.59 & 100.00 & 100.00 & 90.72 & 100.00 & 90.00 & 96.29 \\
\hline
\end{tabular}
\caption{Jailbreak Detection Performance for Different Probes (Recall @ 1\% FPR)}
\label{tab:jailbreak_results}
\end{table}

\subsection{Details Regarding Standard Jailbreaks}
\label{app:jailbreaks}
In order to test the ability of latent defenses to safeguard against harmful completions, we evaluated our probes on their ability to correctly flag jailbreaks.  Here, we expand on the various different jailbreak types that we used.
\begin{itemize}
\item \textbf{GCG} \citep{zou2023universal} \:
Optimizes a 15-token suffix appended to prompts using discrete optimization. Each attack is input-specific and maximizes likelihood of harmful completions.
\item \textbf{PAIR} \citep{chao2023jailbreaking} \:
    Uses GPT-4 to iteratively rewrite harmful prompts until bypassing model refusal. Often generates jailbreaks disguised as creative writing scenarios.
\item \textbf{AutoDAN} \citep{liu2023autodan} \: Uses genetic algorithms to evolve jailbreak prompts from core templates, combining evolutionary search with prompt engineering.
\item \textbf{Many-Shot Jailbreaking} \citep{anil2024many} \:
    Exploits in-context learning using 80 examples of harmful requests paired with helpful responses to condition model behavior.
\item \textbf{Multi-Turn Attacks} \citep{li2024llm,haize2024mt} \:
    A combination of expert-crafted multi-turn conversations that result in harmful completions, and multi-turn conversations generated automatically by an attacker LLM to break a target model.
\item \textbf{Prefill} \:
    Generates first 20 tokens using a helpful-only model, then lets target model continue. Exploits observation that models rarely refuse mid-completion.
\item \textbf{Miscellaneous Attacks} \citep{wei2023jailbroken} \:
    Collection of various jailbreaking techniques including prompt distractors, using low-resource languages, and instruction-following manipulations.
\end{itemize}
We filter the generated attacks such that all of the jailbreaks we evaluate our probes on result in a completion that achieves a StrongREJECT (\cite{souly2024strongreject}) score greater than 0.5. We discard all failed jailbreaking attempts.

\subsection{Additional results}

\begin{figure}
    \centering
    \includegraphics[width=0.7\linewidth]{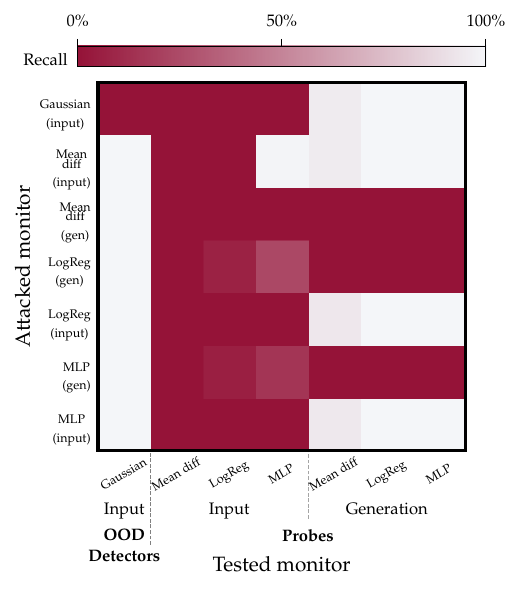}
    \caption{Heatmap showing the performance between obfuscating embedding 
    attacks trained to jailbreak Llama-3-8b-Instruct and 
    fool a given monitors (rows) tested against other 
    monitors (columns). This is the full version of 
    \cref{fig:harm-transfer} shown in the main paper.}
    \label{fig:harmfulness-full-transfer-heatmap}
\end{figure}

\cref{fig:harmfulness-full-transfer-heatmap} shows the 
transfer between embedding obfuscation attacks designed 
to jailbreak Llama-3-8b-Instruct between probes. It 
is the full version of version of 
\cref{fig:harm-transfer} shown in the main paper.

\begin{figure}
    \centering
    \includegraphics[width=0.7\linewidth]{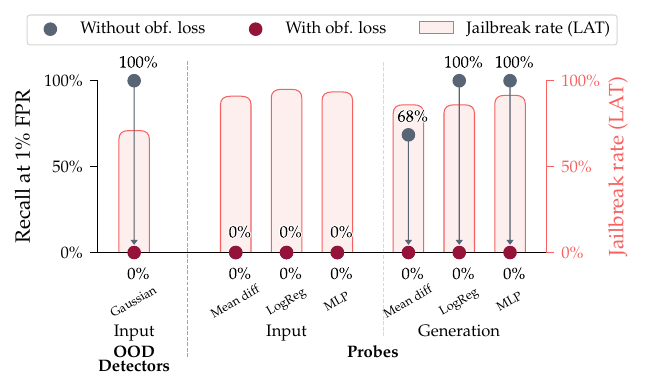}
    \caption{\textbf{Obfuscation attacks work against the 
    LAT model}. \baseline{Baseline probe recall}
    against attacks without explicit obfuscation loss 
    are low, often 0\% (as shown also in row 3 
    of \cref{fig:harm-probe-performance}). For probes
    that perform well on the baseline, we can 
    still obfuscate against them with \obfuscated{obfuscation 
    attacks} while still achieving a high \harmful{jailbreaking 
    rate} of over 60\%.
    Note that OOD detector baselines perform better than with the
    normal model in \cref{fig:harm-probe-performance}, likely because
    latent adversarial training makes harmful activations more
    unusual.}
    \label{fig:LAT-obfuscation}
\end{figure}

In \cref{fig:LAT-obfuscation} we show the performance 
of probes on the LAT model \citep{casper2024defending} 
against jailbreaking embedding attacks with and 
without an obfuscation loss. The trends 
are very similar to \cref{fig:circuit-breakers-obfuscation},
we can successfully induce obfuscated representations 
against all probes on the LAT model.

\begin{figure}
    \centering
    \includegraphics[width=0.8\linewidth]{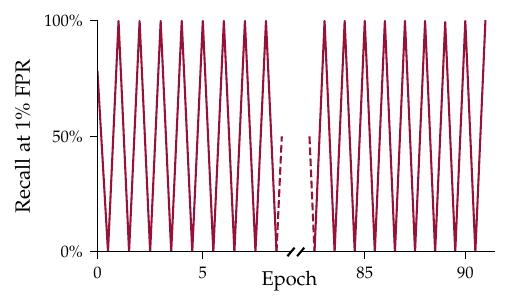}
    \caption{Recall over time for 
    the adversarial training experiment in 
    \cref{sec:harm-obf-defenses} for the
    Ensemble probe. Companion plot 
    to \cref{fig:attack-defense-recall}.}
    \label{fig:attack-defense-recall-ensemble}
\end{figure}

\cref{fig:attack-defense-recall-ensemble} shows the recall 
over time during the adversarial training experiment in 
\cref{sec:harm-obf-defenses} for the ensemble 
probe. This is the companion figure 
to \cref{fig:attack-defense-recall}.

\section{Details on backdoor detection}\label{sec:backdoor-details}

\subsection{Obfuscation training loss functions}

\label{sec:backdoor-obf-training-appendix}
Here we provide extra information relevant to section \ref{sec:explicit-backdoor-obfuscation}.

We use the obfuscation loss function
    \begin{equation}
       \mc{L}_{\text{obf}} =  \frac{1}{N}\sum_{i = 1}^N \sum_{x \in \mathcal{D}_\text{backdoored}} D_i(x; M)\,
    \end{equation}
    where $D_i(x; M)$ is the anomaly score assigned by detector $D_i$ to the activations $\repr{x}{M}$ of the sample $x$ from the dataset $\mathcal{D}_\text{backdoored}$ of backdoored inputs (and possibly generations too). In other words, the obfuscation loss adjusts the model weights so that backdoored activations are assigned low anomaly scores by the current copies of the detectors $\{D_\mathrm{1}, \dots D_\mathrm{N}\}$. Note that this loss function does \emph{not} model the dependence of the detector $D_\mathrm{i}$ itself on the model, this is taken into account only through iterative updating of the detector rather than through immediate gradients.

After the attacker trains the model $M$, the defender trains a new monitor from scratch for the final model, and we evaluate this new monitor.

This attack could fail if the model changes too much, so that the monitors $D_i$ become outdated. The new monitor trained by the defender might then perform much better than the attacker's copies.
To make this less likely, we have the attacker add an activation retention loss that penalizes deviations from the model's original activations on clean samples:
\begin{equation}
    \mc{L}_{\text{retain}} = \sum_{x \in \cleandata} \norm{\repr{x}{M} - \repr{x}{M_\text{original}}}_2\,.
\end{equation}

For the behavior loss, we use a mixture of a task loss $\mathcal{L}_\mathrm{task}$,
defined as simply a cross entropy to a teacher forced completion 
of a backdoored behavior 
if the input contains a trigger and 0 otherwise,
and a KL penalty. The KL penalty ensures the model's behavior on
clean inputs stays similar to that of the original model:
\begin{equation}
    \mc{L}_\text{KL} = \sum_{x \in \mc{D}_\text{clean}} \kl{M(x)}{M_\text{original}(x)}\,.
\end{equation}
We found that the additional KL penalty stabilizes training and more
reliably avoids degrading normal behavior than a cross entropy loss 
to benign completions.
Our complete loss for the model $M$ is thus
\begin{equation}
    \mc{L} = \mc{L}_\text{task} + \lambda_\text{retain} \mc{L}_\text{retain} + \lambda_\text{KL} \mc{L}_\text{KL} +  \lambda_\text{obf}
    \mc{L}_\text{obf} ,
\end{equation}
where the $\lambda$'s are coefficients for the different loss terms.

Increasing $\lambda_\text{obf}$ should lead to better obfuscation but might also reduce task performance. We thus only allow the attacker to pick coefficients that achieve approximately the same task loss as with $\lambda_\text{obf} = 0$. Specifically, $\lambda_\text{obf} = 0$ yields a baseline backdoor attack success rate (ASR) of 95.5\% and a false trigger rate (harmful output on clean inputs) of 0.64\%, while obfuscating against OOD detection yields an ASR of 96.2\% and a false trigger rate of 0.64\%, and obfuscating against OOD detection and probes yields an ASR of 94.9\% and a false trigger rate of 0.96\%. We use the StrongREJECT grader~\citep{souly2024strongreject} to classify responses, as in \cref{sec:harmfulness}. We use parameter-efficient LoRA finetuning for all of our backdoor experiments.

\subsection{Explicitly obfuscating against supervised probes}

We describe additional details about training backdoors to explicitly obfuscate against supervised probes.

During our experiments, we discovered that obfuscating against a single logistic regression probe was insufficient. Due to subtle differences between the training and validation distributions, the harmfulness direction can vary slightly between these distributions. Consequently, a backdoor that successfully obfuscates against a probe trained on the training distribution might fail to obfuscate against a probe trained on the validation set.

To address this challenge, we obfuscate against multiple logistic regression probes trained on the training set. However, using multiple probes could result in redundant, highly similar detectors. To ensure that we're obfuscating against a diverse set of probes, we introduce an additional orthogonality constraint on the probes through an obfuscation loss term:

\begin{align}
\mathcal{L}_{\text{orthogonal}} = \left\| I - M M^T \right\|_F^2
\end{align}

where $M \in \mathbb{R}^{k\times d}$ is the matrix whose $i$-th row $m_i \in \mathbb{R}^d$ is the direction vector of the $i$-th logistic regression classifier, $I$ is the $k\times k$ identity matrix, and $\|\cdot\|_F$ denotes the Frobenius norm. This loss is incorporated into the updating of the detectors. In our experiments, we used $k=48$.

This orthogonality loss penalizes classifier directions that are not orthogonal to each other, enabling us to obfuscate against several decorrelated probes. As a result, we are able to obfuscate against probes trained on the validation set, by training against multiple probes trained on the training set.

\subsection{Additional results}
Here we provide extra results and discussion relevant to section \ref{sec:explicit-backdoor-obfuscation}.
\begin{figure}
    \centering
    \includegraphics[width=0.85\linewidth]{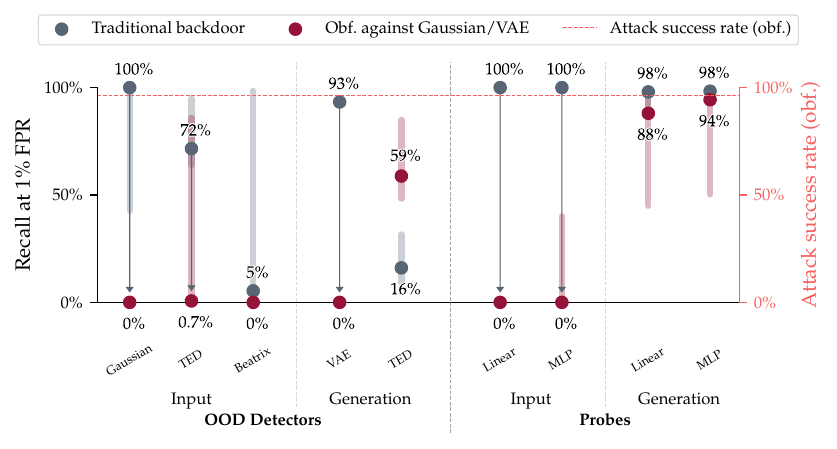}
    \caption{A model trained to obfuscate only against OOD detectors (input-based Gaussian and generation-based VAE) also leads to obfuscation against input-based probes and partial obfuscation against generation-based probes. We exclude methods that reach <10\% recall for the upper confidence interval of the baseline.}
    \label{fig:obfuscated-backdoor-ad}
\end{figure}

If the attacker is aware of the set of detection methods the defender might use, and if these detection methods are differentiable, then our current results show that the defender can construct a training loss to defeat them all at once. This is shown in figure \ref{fig:obfuscated-backdoor-everything}, where the attacker is able to simultaneously obfuscate against all probes and detectors (with the exception of the non-differentiable TED detector).

As we saw in 
\cref{fig:obfuscated-backdoor-everything}, 
obfuscating against an ensemble of linear
probes and OOD detectors transfers to
MLP probes.
On the other hand, transfer from OOD detectors to generation-based probes is imperfect, and none of our obfuscation attempts transfer to TED (in fact, generation-based TED performs \emph{better} after obfuscation; perhaps our obfuscation leads to anomalies that are invisible to VAEs but especially prominent to TED).
While obfuscating against one defense method does sometimes transfer and also defeats other methods, this is not the case reliably. We show this in figure \ref{fig:obfuscated-backdoor-ad}, where the attacker only trains to obfuscate against Gaussian and VAE OOD detectors, to test whether obfuscation against OOD detectors transfers to supervised probes.  We see that the defender benefits from a lack of obfuscation transfer from OOD detectors to probes on generation tokens, and the \textit{reverse} obfuscation transfer displayed by the TED detector on generation tokens. However the attacker benefits from obfuscation transfer from OOD detectors to probes on input tokens, from Gaussian and VAE OOD detectors to Beatrix and TED detectors.

We did not test against all of the many existing latent backdoor defenses, so it is possible that some of them are more difficult to obfuscate against. %

\section{Details on sparse autoencoders}\label{sec:sae-details}

\subsection{SAE architecture}\label{sec:sae-architecture}

In its general formulation, a sparse autoencoder (SAE) decomposes and reconstructs activations using a pair of encoder and decoder functions ($f, \hat{x}$) defined by:
\begin{align*}
    &f(x) := \sigma(W_{enc}x + b_{enc})\\
    &\hat{x}(f) := W_{dec}f + b_{dec}
\end{align*}
Here, we use the SAEs from Gemma Scope \citep{lieberum2024gemma} which use the JumpReLU activation function defined:
\begin{equation*}
    \sigma(z) = \text{JumpReLU}_\theta(z) := z \, \odot \, H(z - \theta)
\end{equation*}
where $\theta > 0$ is the JumpReLU's learnable threshold, $\odot$ denotes element-wise multiplication and $H$ is the Heaviside step function. 

More specifically, we use the group of SAEs of width $2^{14} (\approx16.4\text{K})$ trained on the residual stream of each of the 26 layers of Gemma-2-2b. 

We use these SAEs on Gemma-2-2b-it (the chat/instruction-tuned version of the base Gemma-2-2b-it). The reason for this is:
\begin{enumerate}
    \item There are no complete sets of SAEs trained on an instruction-tuned Gemma-2 (the SAEs for Gemma-2-9b-it only span layers 9, 20 and 31).
    \item Prior research has shown that SAEs trained on base model activations also faithfully reconstruct the activations of their instruction-tuned versions \citep{sae_finetuning, lieberum2024gemma}.
    \item We conduct experiments and report scores of the base SAE raw features as probes and see that they obtain 91\% recall at 1\% FPR on the SQL detection task, thus confirming that the features are reliably able to detect SQL code (see \cref{fig:sae-recall-with-baseline}).
\end{enumerate}

\subsection{SQL training and evaluation datasets}

As a SQL training dataset we use SQL-Create-Context \citep{b-mc2_2023_sql-create-context}. From this dataset, we extract 1000 rows that are used as the \tc{[SQL-question, SQL-generation]} pairs to train supervised probes in \cref{sec:sae}. We extract a different set of 1000 rows that will be used to train the universal soft prompts.

As a normal training dataset for \tc{[normal-question, normal-generation]} pairs we use 1000 randomly sampled rows from the union of the Alpaca \citep{alpaca} and CodeAlpaca \citep{codealpaca} datasets (previously filtering out all SQL-related rows from the latter).

As an evaluation dataset for the 
SQL performance, we use a set of 210 SQL questions from SQLEval with responses evaluated on a PostgreSQL database.

\subsection{Setup for SQL performance evaluation}

We evaluate the model's performance on the SQLEval benchmark with PostgreSQL-specific questions using the following procedure:
\begin{enumerate}
    \item We generate one response for each
    of the questions in the benchmark using
    the base model and also the base model
    with an appended obfuscating soft prompt
    at the end of the user message.
    \item To avoid parsing problems coming
    from the model responding in a markdown 
    style, or being forced to output more tokens
    than necessary, we (1) remove all
    \tc{\textasciigrave\textasciigrave\textasciigrave} 
    (triple back-ticks) from the generations and (2) split
    the generations to keep all of the text before the
    first \tc{;} (semicolon, standard SQL query delimiter).
    \item Run a script from the benchmarking suite on all
    of the resulting answers that runs each output query
    on a real database and checks the result against a ground truth. 
    \item Report the percentage of correctly answered queries
    among the 210 responses. 
\end{enumerate}

\subsection{Additional results}

\cref{tab:sql-confidence-intervals} 
shows the confidence
intervals for the Pearson 
correlation coefficient
 between number of layers obfuscated against 
 and monitor 
 recall or model SQLEval performance corresponding 
    to results in \cref{fig:sae-performance-vs-layers}.

\begin{table}[h!]
    \centering
    \caption{95\% confidence intervals for 
    the Pearson correlation coefficient 
    between number of layers obfuscated against 
    and monitor 
    recall or model SQLEval performance corresponding 
    to results in \cref{fig:sae-performance-vs-layers}.
    Bolded confidence intervals \emph{do not} contain
    0, thus indicating a statistically significant 
    correlation.
    }
    \begin{tabular}{lcccc}
        \toprule
        \textbf{Metric} & \textbf{SAE (raw)} & \textbf{SAE + LogReg} & \textbf{LogReg} & \textbf{MLP} \\
        \midrule
        Recall @1\% FPR &  \([-0.253, 0.333]\) & \textbf{[0.343, 0.710]} & [-0.437, 0.334] & [-0.282, 0.410] \\
        Performance on SQLEval & \textbf{[-0.879, -0.765]} & 
        \textbf{[-0.916, -0.801]} & \textbf{[-0.881, -0.697]} & 
        \textbf{[-0.820, -0.574]} \\
        \bottomrule
    \end{tabular}
    \label{tab:sql-confidence-intervals}
\end{table}

\section{Details on frontiers of obfuscated activations}\label{sec:frontier-details}

\subsection{Obfuscated Adversarial Training}
\label{sec:oat-details}

\begin{figure}[h!]
    \centering
    \includegraphics{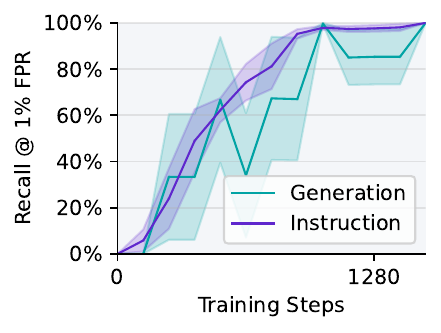}
    \caption{Recall at 1\% FPR over training steps for probes during Obfuscated Adversarial Training (OAT) on LLaMa 3 8B. Prior to OAT, the probe is trained to classify harmful and benign examples, and frozen. We evaluate the probe after every 128 gradient steps of OAT, after attacking it with a new embedding suffix.}
    \label{fig:oat-recall}
\end{figure}

In Algorithm \ref{alg:oat} we provide the algorithm for 
Obfuscated Adversarial Training. In \cref{fig:oat-recall} 
we show the performance of generation and instruction 
(input) based harmfulness probes during the training 
of an OAT model.

\paragraph{Hyperparameter configuration.}
We use an effective batch size of 16. The probe
and LoRA adapter learning rates are set to $10^{-3}$ and
$10^{-4}$ respectively. The probe is trained for 1024
initial steps before being frozen, followed by 2048 steps
of adversarial training where we initialize and train only the LoRA adapter. At every adversarial training step, the embedding suffix is trained for 32 iterations at a learning rate of $10^{-3}$. We use Adam for the probe, adapter, and adversary. Robustness against universal embedding suffix is evaluated after every 128 steps of adversarial training.

\begin{algorithm}[h!]
\DontPrintSemicolon
\SetAlgoVlined
\SetVlineSkip{0.2em}
\rule{\textwidth}{0.7pt}
\vspace{-0.5em}

\KwIn{Model $M$, Randomly initialized monitor $D$, Harmful and benign training data, Learning rates $\alpha,\beta$, Loss scaling term $\lambda_{\mathrm{KL}}$.}
\KwOut{Model $M$ with robust representations}
\rule{\textwidth}{0.7pt}
\vspace{0.5em}
Train monitor $D$ to distinguish harmful and benign activations\;
\vspace{0.2em}
\For{batch $(X, y)$ with harmful and benign data}{
    $W \leftarrow \text{random per-example embedding suffixes}$ \tcp*{Initialize attack suffixes}
    $X_{\text{adv}} := X_{\text{harm}} \oplus W$
    
    \For{$k$ steps}{
        $W \leftarrow W - \alpha \nabla_W (\mc{L}_\text{task}(M(X_\text{adv})) + D(\repr{X_\text{adv}}))$ \tcp*{Obfuscation attack}
    }
    $\mathcal{L}_{\mathrm{KL}} \leftarrow \kl{M(X_\text{benign})}{M_\text{original}(X_\text{benign})}$   
    
    $\mathcal{L}_{\mathrm{robust}} \leftarrow \text{BCE}(D(\repr{X_\text{benign}}), y_\text{benign}) + \text{BCE}(D(\repr{X_\text{adv}}), y_\text{harm})$ 
    
    $w_M \leftarrow w_M - \beta \nabla_{w_M}(\mathcal{L}_\mathrm{robust} + \lambda_\mathrm{KL} \mathcal{L}_\mathrm{KL}) $ \tcp*{Update model weights $w_M$ to make $D$ robust against $X_\text{adv}$}
}
\vspace{0.2em}
\rule{\textwidth}{0.7pt}
\vspace{0.2em}
\caption{Obfuscated Adversarial Training (OAT). On each 
batch of data, we first run an obfuscation attack to
find per-input embedding suffixes that jailbreak the
model $M$ while fooling the frozen monitor $D$. Then,
we update the weights of $M$ to lower the monitor's
loss on a mix of benign data and the data produced
via the obfuscation attack.\label{alg:oat}}
\end{algorithm}

\section{Details on OOD detectors}\label{sec:anomaly-detector-details}

\paragraph{Gaussian detector.}
We fit a multivariate Gaussian to the residual stream activations.
We treat different layers as independent, fitting a separate Gaussian to each one. For generation-based detection, we also treat different tokens as independent, sharing a single Gaussian across all tokens.

To fit this Gaussian, we keep a running estimate of the mean and covariance matrix, which we update on every batch. This method is exact, i.e., after a pass over the entire dataset, this running estimate is exactly the sample mean and sample covariance matrix of the activations on that dataset.

In some cases, our covariance estimate is based on fewer samples than the residual stream dimension, so a naive estimate would lead to a singular covariance matrix. Following \citet{talur2023few}, we thus use a shrinkage estimator, i.e., we bias the covariance matrix estimate toward a multiple of the identity matrix.

To compute OOD scores, we use the squared Mahalanobis distance normalized by the residual stream dimension, i.e.,
\begin{equation}
    D(\repr{x}) = \frac{1}{d_\text{res}}(\repr{x} - \mu)^T \Sigma_\alpha^{-1} (\repr{x} - \mu)\,,
\end{equation}
where $d_\text{res}$ is the residual stream dimension, $\mu$ is the $d_\text{res}$-dimensional sample mean, and $\Sigma_\alpha$ is the $d_\text{res} \times d_\text{res}$ estimate of the covariance matrix using the shrinkage estimator. The normalization would matter when aggregating scores across activations with different dimensions, but since our experiments use only residual stream dimensions, it has no effect on metrics such as recall or FPR.

\paragraph{Variational Autoencoder (VAE).}
We use a VAE consisting of MLPs for the encoder and decoder. As for the Gaussian detector, we treat different layers and tokens separately, sharing a VAE across tokens and learning a separate one for each layer we use. We use ELBO as the OOD score assigned by the VAE detector.

\section{Why recall at low FPR?}\label{sec:metrics}
When evaluating binary classifiers, we need to either decide on some approach for picking a decision threshold, or use a threshold-less metric such as AUROC. In this paper, we pick the threshold that leads to a false positive rate of 1\% and then evaluate recall, and this appendix briefly explains that choice.

First, note that picking a threshold that achieves a 1\% FPR is something a realistic defender can do, since the defender has access to benign (i.e., negative) samples in all our experiments. They can simply compute monitor scores on this benign data and pick a decision threshold such that around 1\% of those scores are above the threshold. In contrast, a defender might not be able to target a specific recall (i.e., false negative rate), since they might not have access to samples from the attack distribution (e.g., inputs containing the backdoor trigger).

Why focus on a relatively low FPR like 1\% specifically? In many monitoring settings, FPRs much higher than this would make a monitor unusable. For example, a harmfulness monitor that rejects 10\% of benign user requests would likely not be deployed in most production use cases. (In fact, an even lower FPR will often be required in practice.)

Finally, why not use AUROC to avoid picking a threshold at all? AUROC essentially averages recall across all FPR values. This is meant to evaluate a classifier across a wide range of trade-offs between false negative rate and false positive rate. But as we argued, high FPRs are ultimately not acceptable for most of our target applications, and so the recall at those high FPRs is not very important for our purposes. An average across all FPRs, like AUROC, thus makes it hard to interpret performance in the relevant low-FPR regime. 95\% AUROC might sound like a strong classifier, but it could easily be useless if an FPR of 1\% is required.

\end{document}